\documentclass[logo, address]{trfr} 

\usepackage{natbib}

\usepackage{hyperref}
\usepackage{amsmath}

\usepackage{booktabs}  
\usepackage{multirow}  
\usepackage{graphicx}  
\usepackage{array}     
\usepackage{xcolor}    
\usepackage{adjustbox} 

\usepackage{wrapfig}
\usepackage{xcolor} 

\usepackage{tabularx}
\usepackage{makecell}
\usepackage{ragged2e}
\usepackage[table]{xcolor}

\usepackage{makecell}
\usepackage{pifont}
\newcommand{\cmark}{\ding{51}}

\definecolor{canColor}{HTML}{D64000}   
\definecolor{willColor}{HTML}{4DB299}  
\definecolor{howColor}{HTML}{E9B045}   
\definecolor{takeawaysColor}{HTML}{233454}

\title{CapTrack: Multifaceted Evaluation of Forgetting in LLM Post-Training}

\abstract{
    Large language model (LLM) post-training enhances latent skills, unlocks value alignment, improves performance, and enables domain adaptation. 
    Unfortunately, post-training is known to induce forgetting, especially in the ubiquitous use-case of leveraging third-party pre-trained models, which is typically understood as a loss of parametric or factual knowledge. We argue that this accuracy-centric view is insufficient for modern foundation models and instead define forgetting as systematic model drift that degrades behavior and user experience.
    In this context, we introduce \textbf{CapTrack}, a capability-centric framework for analyzing forgetting in LLMs that combines a behavioral taxonomy with an evaluation suite centered on capability-specific metrics.
    Using CapTrack, we conduct a large-scale empirical study across post-training algorithms, domains, and model families, including models up to 80B parameters.
    We find that forgetting extends beyond parametric knowledge, with pronounced drift in robustness and default behaviors. Instruction fine-tuning induces the strongest relative drift, while preference optimization is more conservative and can partially recover lost capabilities. Differences across model families persist, and no universal mitigation emerges.
}

\author[1,2,3]{Lukas Thede}
\author[1]{Stefan Winzeck}
\author[3,4]{Zeynep Akata}
\author[1,5]{Jonathan Richard Schwarz}

\affiliation[1]{Thomson Reuters Foundational Research}
\affiliation[2]{Tübingen AI Center, University of Tübingen} 
\affiliation[3]{Helmholtz Munich}
\affiliation[4]{Munich Center for Machine Learning (MCML), Technical University Munich}
\affiliation[5]{Imperial College London}

\correspondence{\{first.last\}@thomsonreuters.com}

\begin{document}

\maketitle

\section{Introduction}
Adapting large language models (LLMs) through post-training is central to enhancing latent skills, improving performance, unlocking value alignment, and enabling adaptation to new application settings. 
While effective, post-training is known to induce forgetting, typically understood as the loss of previously acquired parametric knowledge. Prior work has therefore studied forgetting mainly through declines in factual accuracy or task performance. Although informative, this captures only a narrow notion of model quality. In early continual learning (i.e. vision) this focus was appropriate as models were optimized for a single performance metric. In contrast, modern LLMs are interactive systems whose usefulness depends on a broader set of capabilities, including reasoning, robustness, default behavioral preferences, and adherence to interaction protocols.

For instance, a Qwen 3 80B Next model post-trained on legal data may appear stable on standard benchmarks, with negligible changes in factual accuracy ($+0.4\%$ on MMLU-Pro) and instruction-following scores ($+0.7\%$ on IFEval), yet exhibits substantial degradation in other capabilities such as multilingual robustness ($-16.4\%$) and verbosity ($-4.3\%$). Thus, evaluating forgetting solely through parametric knowledge risks missing model drift that directly affects model behavior and user experience. We therefore extend the notion of forgetting to encompass any systematic drift that harms capabilities, behaviors, or execution reliability.

To measure all aspects captured by the wider definition, we introduce \textbf{CapTrack}, a capability-centric framework for assessing forgetting in LLM post-training. CapTrack combines a structured taxonomy that decomposes LLM behavior into latent competence, default behavioral preferences, and protocol compliance with an evaluation suite that probes these dimensions using established benchmarks with targeted adaptations. We argue that designing a new comprehensive benchmark is undesired and impractical, given the substantial interdisciplinary effort required and the rapid pace at which benchmarks become saturated and obsolete. Instead, by mapping existing benchmarks onto a capability spectrum, CapTrack serves as meta-benchmark framework, offering multifaceted and cumulative insights into post-training effects and adaptability through the integration of new benchmarks over time.

\begin{wrapfigure}{r}{0.52\columnwidth} 
  \centering  
  \vspace{-6pt} 
  \includegraphics[width=0.51\columnwidth]{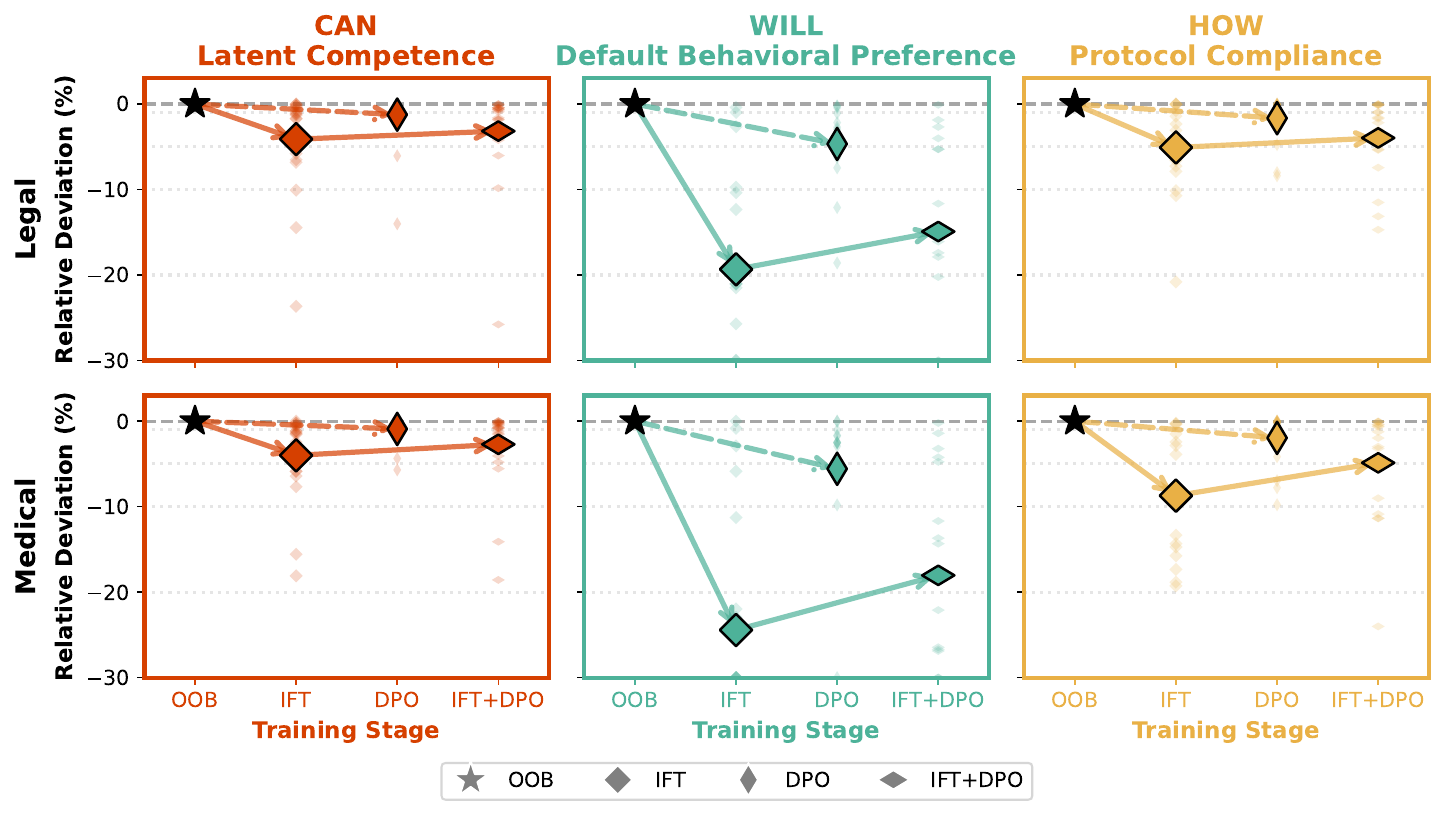}  
  \vspace{-4pt}  
  \caption{Average forgetting across post-training stages (relative to OOB; higher is better).  
  Forgetting extends beyond factual knowledge, with strongest degradation under IFT, milder effects under DPO, and partial recovery when DPO is applied after IFT.}  
  \label{fig:forgetting_stage_comparison}  
  \vspace{-10pt} 
\end{wrapfigure}  

We employ CapTrack in a large empirical study across the legal and medical domains, covering 7 models from the Llama, Qwen, and Gemma families with up to 80B parameters. We find that forgetting consistently extends beyond parametric knowledge, most notably in robustness, default behavior preferences, and multi-turn consistency.
Patterns differ systematically across model families but show no consistent dependence on model size. Instruction fine-tuning (IFT) induces the strongest degradation, whereas direct preference optimization (DPO) is more conservative and can partially recover lost capabilities.

Finally, we use CapTrack to assess several mitigation strategies at the capability level. Forgetting is not tied to domain-specific adaptation: data interventions yield mixed, capability-dependent effects, while global regularization reduces forgetting only by sacrificing post-training gains, revealing a fundamental stability-plasticity trade-off. 

Overall, our results show that forgetting in LLM post-training is fundamentally a capability-level phenomenon that cannot be assessed by retention of general factual knowledge alone. Understanding and mitigating it, therefore, requires moving beyond traditional general knowledge benchmarks toward a holistic, capability-centric evaluation of model behavior.

These findings motivate a more structured examination of post-training effects beyond task-level accuracy. CapTrack provides such a framework by combining targeted benchmark adaptations with capability-specific measurement protocols and a unified relative-evaluation methodology, enabling consistent analysis of capability drift across models, domains, and post-training methods.

\begin{itemize}
    \item We extend the notion of forgetting in LLM post-training beyond parametric knowledge loss to encompass capability-level drift in behavior and execution.

    \item We introduce \textbf{CapTrack}, a capability-centric evaluation framework that combines a behavioral taxonomy with targeted benchmark adaptations and capability-specific measurement protocols, enabling decomposition of model behavior beyond task-level metrics.


    \item Using CapTrack, we provide a unified empirical analysis of capability-level forgetting across post-training algorithms, data sources, model families and scales, and various mitigation strategies, revealing capability-specific effects and fundamental stability-plasticity trade-offs.
\end{itemize}

\section{Related Work}

\subsection{Post-Training of Large Language Models}
Post-training has become the primary driver of capability gains in modern LLMs, often outweighing additional pretraining scale in its impact on downstream performance \citep{openai2024openaio1card}, with contemporary pipelines operating at data scales comparable to early pretraining \citep{tie2025surveyposttraininglargelanguage}. Post-training commonly covers supervised fine-tuning on instruction–response pairs (i.e. IFT) \citep{zhang2025instructiontuninglargelanguage}, preference optimization via ranked or pairwise feedback (e.g., DPO \citep{rafailov2024directpreferenceoptimizationlanguage}, IPO \citep{garg2025ipolanguagemodelsecretly}, KTO \citep{ethayarajh2024ktomodelalignmentprospect}), and reinforcement learning–based alignment using learned rewards, most prominently PPO-based RLHF and GRPO \citep{ouyang2022traininglanguagemodelsfollow,shao2024deepseekmathpushinglimitsmathematical}. In this work, we focus on IFT and DPO post-training.

\subsection{Forgetting in Neural Networks and LLMs}
\begin{table*}[t]
\centering
\scriptsize
\setlength{\tabcolsep}{3pt}
\renewcommand{\arraystretch}{1.15}
\resizebox{\textwidth}{!}{%
\begin{tabular}{l c c c c c c c c c c c c c | c}
\toprule
\textbf{Capability} 
& \textbf{L\&L'24} 
& \textbf{Yil'24} 
& \textbf{Jia'24} 
& \textbf{Luo'23} 
& \textbf{Haq'25} 
& \textbf{Hua'24} 
& \textbf{Zhe'25} 
& \textbf{LiuN'25} 
& \textbf{Har'25} 
& \textbf{Fer'24} 
& \textbf{Kot'23} 
& \textbf{Li'25} 
& \textbf{Lin'24} 
& \textbf{CapTrack} \\
\midrule
\textbf{Setting} 
& CPT & CPT & IFT & IFT & IFT & IFT & IFT & IFT & IFT, RL & IFT, DPO & IFT, RL & IFT, RL & RL & IFT, DPO \\
\midrule
\rowcolor[HTML]{F8DED7} Parametric Knowledge          & \cmark & \cmark & \cmark & \cmark & - & \cmark & \cmark & - & \cmark & \cmark & - & - & \cmark & \cmark \\
\rowcolor[HTML]{F8DED7} Reasoning                     & - & - & - & \cmark & - & - & - & - & \cmark & - & - & \cmark & - & \cmark \\
\rowcolor[HTML]{F8DED7} In-Context Learning (ICL)     & - & - & - & \cmark & \cmark & \cmark & - & - & - & - & - & \cmark & \cmark & \cmark \\
\rowcolor[HTML]{F8DED7} Faithfulness                  & - & - & - & - & - & - & - & - & - & - & - & - & - & \cmark \\
\rowcolor[HTML]{F8DED7} Robustness                    & \cmark & - & - & - & - & - & - & \cmark & \cmark & \cmark & - & - & \cmark & \cmark \\
\rowcolor[HTML]{E0F0EC} Willingness to Answer         & - & - & - & - & - & - & - & - & \cmark & - & \cmark & - & - & \cmark \\
\rowcolor[HTML]{E0F0EC} Informational Scope           & - & - & - & - & - & - & - & - & - & - & - & - & - & \cmark \\
\rowcolor[HTML]{E0F0EC} Style \& Elaboration          & - & - & - & - & - & - & - & - & - & - & - & - & - & \cmark \\
\rowcolor[HTML]{FBF0DE} Instruction Following         & - & - & \cmark & - & - & - & - & - & - & - & \cmark & - & - & \cmark \\
\rowcolor[HTML]{FBF0DE} Output-Format Fidelity        & \cmark & - & - & - & - & - & - & - & - & - & - & - & - & \cmark \\
\rowcolor[HTML]{FBF0DE} Tool / Function Use           & - & - & - & - & - & - & - & - & - & - & - & - & - & \cmark \\
\rowcolor[HTML]{FBF0DE} Multi-turn Commitment         & - & - & - & - & - & - & - & - & - & - & - & - & - & \cmark \\
\rowcolor[HTML]{FBF0DE} Context-Window Operations     & - & - & - & - & - & - & - & - & - & - & - & - & - & \cmark \\
\rowcolor[HTML]{FBF0DE} Citation Mechanics            & - & - & - & - & - & - & - & - & - & - & - & - & - & \cmark \\
\bottomrule
\end{tabular}%
}
\caption{
Coverage of capabilities considered in prior studies on forgetting in LLM post-training. Rows denote capability categories and columns individual studies; the setting row indicates the post-training regime. Existing work focuses primarily on parametric knowledge, whereas CapTrack provides comprehensive coverage across competence, behavior, and execution capabilities.
}
\vspace{-15pt}
\label{tab:forgetting_literature_inverted}
\end{table*}
Catastrophic forgetting, the degradation of previously acquired knowledge during training on new data, has been widely studied in continual learning, primarily in sequential task settings and with a focus on parametric knowledge preservation \citep{mccloskey1989catastrophic,kirkpatrick2017overcoming,parisi2019continual}. Classical approaches mitigate forgetting via regularization, replay, or architectural constraints, typically evaluated through task- or class-level accuracy.

More recently, forgetting has been examined in LLMs under continued pre- and post-training regimes. Several works study forgetting during continual pre-training, focusing on the retention of factual knowledge \citep{li2024revisiting,Yildiz2024-pw}. Multiple post-training studies analyze forgetting induced by instruction fine-tuning or alignment, again primarily through changes in parametric knowledge \citep{Jiang2024-ey,Luo2023-lw,haque2025catastrophicforgettingllmscomparative,Huang2024-ac}. Extensions beyond factual recall are limited: some works consider robustness to distribution shift or multilingual settings \citep{liu2025conditionscatastrophicforgettingmultilingual,harmon2025mappingposttrainingforgettinglanguage}, while others analyze forgetting in specific capabilities such as reasoning or instruction following \citep{Luo2023-lw,Kotha2023-lr,Li2025-hq}.

A smaller number of studies investigate alignment- and preference-related effects, including forgetting under reinforcement learning or preference optimization \citep{Fernando2024-yx,lin2024mitigatingalignmenttaxrlhf}. However, these analyses typically focus on a narrow subset of behaviors (e.g., refusal or safety compliance) and do not examine how such changes interact with other aspects of model behavior.

Table~\ref{tab:forgetting_literature_inverted} highlights that most prior work evaluates only a small fraction of the capabilities relevant to real-world LLM behavior. In particular, behavioral and interaction-level capabilities remain largely unexplored. This gap motivates a holistic, capability-centric framework for analyzing forgetting in LLM post-training, which we address with CapTrack.

While some prior work evaluates non-factual capabilities (e.g., robustness or reasoning), these are typically studied in isolation as task-level metrics. In contrast, CapTrack groups capabilities into interpretable dimensions and evaluates them under a unified relative-deviation framework, enabling analysis of interactions and trade-offs.

\section{CapTrack: Capability-Level Forgetting}
\label{sec:captrack}

\begin{table}[t!]  
\centering  
\small  
\setlength{\tabcolsep}{6pt}  
\renewcommand{\arraystretch}{1.1}  
\resizebox{1\textwidth}{!}{%
\begin{tabular}{l l l}  
\toprule  
\textbf{Capability} & \textbf{Benchmarks} & \textbf{Metrics} \\  
\midrule

\rowcolor[HTML]{D64000}  
\textbf{\textcolor{white}{\textbf{CAN – Latent Competence}}}%
& \textcolor{white}{\textit{What the model can do if prompted ideally}}
& \\  

\rowcolor[HTML]{F8DED7}  
\textbf{C1:} Parametric knowledge \& Skills
& MMLU-Pro \citep{wang2024mmluprorobustchallengingmultitask}; PopQA \citep{mallen2023llm_memorization}; GSM8K \citep{cobbe2021gsm8k};  
& Accuracy \\  
\rowcolor[HTML]{F8DED7}  
& LiveMathBench \citep{liu2024your}; HumanEval \citep{chen2021evaluating}; MBPP \citep{austin2021program} & \\

\rowcolor[HTML]{F8DED7}  
\textbf{C2:} Reasoning \& Problem solving
& MATH \citep{hendrycks2021measuringmathematicalproblemsolving}; SuperGPQA \citep{pteam2025supergpqascalingllmevaluation} 
& Acc.; Reasoning Quality; No. Steps \\

\rowcolor[HTML]{F8DED7}  
\textbf{C3:} Contextual comprehension
& HotpotQA \citep{yang2018hotpotqa}; BoolQ \citep{clark2019boolq} 
& Acc.; Evidence Hit \\

\rowcolor[HTML]{F8DED7}  
\textbf{C4:} Epistemic faithfulness \& Grounding
& RAGTruth \citep{niu-etal-2024-ragtruth}; TruthfulQA \citep{lin2022truthfulqameasuringmodelsmimic}
& Hallucination Rate, Acc. \\

\rowcolor[HTML]{F8DED7}  
\multicolumn{3}{l}{\textbf{C5:} Robustness of Competence} \\

\rowcolor[HTML]{F8DED7}  
\hspace{1.25em}C5a: Prompt-form invariance  
& Rephrased MMLU/GSM8K;  
& Accuracy \\

\rowcolor[HTML]{F8DED7}  
\hspace{1.25em}C5b: Domain shift robustness  
& WinoGrande \citep{ai2:winogrande}; HellaSwag \citep{zellers2019hellaswag};  
& Accuracy \\

\rowcolor[HTML]{F8DED7}  
\hspace{1.25em}C5c: Multilingual stability  
& MGSM \citep{shi2022language}; XTREME \citep{hu2020xtreme}
& Accuracy \\

\midrule

\rowcolor[HTML]{4DB299}  
\textbf{\textcolor{white}{\textbf{WILL – Policy \& Behavioral Preferences}}}%
& \textcolor{white}{\textit{What the model will do by default}}%
& \\

\rowcolor[HTML]{E0F0EC}  
\textbf{W1:} Willingness to answer
& GSM8K (benign); HarmBench (unsafe) \citep{mazeika2024harmbench};  
& Compliance/Refusal Rate \\

\rowcolor[HTML]{E0F0EC}  
& RULER (4k) (incomplete) \citep{hsieh2024ruler} & \\

\rowcolor[HTML]{E0F0EC}  
\textbf{W2:} Helpfulness \& Informational scope  
& RAGTruth \citep{niu-etal-2024-ragtruth}; ELI5 \citep{fan2019eli5}
& Coverage; Overreach \\

\rowcolor[HTML]{E0F0EC}  
\textbf{W3:} Style \& Level of elaboration 
& MT-Bench (turn 1) \citep{zheng2023judging}; OASST1 \citep{köpf2023openassistantconversationsdemocratizing}
& Verbosity; Hedging; Formatting \\

\midrule

\rowcolor[HTML]{E9B045}  
\textbf{\textcolor{white}{\textbf{HOW – Protocol Compliance \& Execution}}}%
& \textcolor{white}{\textit{How the model executes instructions}}%
&
\\

\rowcolor[HTML]{FBF0DE}  
\textbf{H1:} Instruction following  
& IFEval \citep{zhou2023instructionfollowingevaluationlargelanguage}; FollowBench \citep{jiang2023followbench}
& Pass Rate; Constraint Satisfaction \\

\rowcolor[HTML]{FBF0DE}  
\textbf{H2:} Output-format fidelity 
& Schema-wrapped MMLU-Pro; GSM8K  
& Parse Rate \\

\rowcolor[HTML]{FBF0DE}  
\textbf{H3:} Tool/function use \& integration  
& BFCL \citep{berkeley-function-calling-leaderboard}; MNMS  \citep{ma2024mms}
& Tool Selection; Argument Acc. \\

\rowcolor[HTML]{FBF0DE}  
\textbf{H4:} Multi-turn state \& commitment keeping
& MT-Bench (turn 2); StructFlowBench \citep{li2025structflowbench}  
& Multi-turn Score \\

\rowcolor[HTML]{FBF0DE}  
\textbf{H5:} Context-window operations
& RULER (32k) \citep{hsieh2024ruler}, LongBench-V2  \citep{bai2024longbench2}
& Accuracy \\

\rowcolor[HTML]{FBF0DE}  
\textbf{H6:} Citation \& Attribution mechanics  
& HotpotQA; QASPER \citep{Dasigi2021ADO}  
& Format Correctness, Source Acc. \\

\bottomrule  
\end{tabular}  
}  
\caption{  
Overview of the CapTrack taxonomy and evaluation suite. CapTrack organizes model capabilities into three complementary groups: \textbf{CAN} (latent competence under ideal prompting), \textbf{WILL} (default behavioral preferences), and \textbf{HOW} (protocol compliance and execution). The taxonomy is designed to isolate distinct but interrelated aspects of model behavior, enabling fine-grained analysis of post-training forgetting beyond parametric knowledge.
Each capability is assessed using established benchmarks and targeted metrics; when automated scoring is insufficient, LLM-as-a-judge evaluation is used (see Appendix~\ref{app:judge} for details).  
}  
\label{tab:captrack_eval_suite}  
\vspace{-10pt}  
\end{table}  

To assess forgetting in LLM post-training beyond parametric knowledge loss, we introduce \textbf{CapTrack}, a capability-centric framework for tracking model drift. CapTrack is motivated by the observation that post-training can alter model behavior in ways not captured by general knowledge benchmarks, yet directly affect user experience. It combines a structured taxonomy of LLM capabilities with an evaluation suite that operationalizes these dimensions through established benchmarks augmented with targeted adaptations and capability-specific metrics. An overview of the CapTrack taxonomy and evaluation suite is provided in Table~\ref{tab:captrack_eval_suite}. Together, CapTrack enables a multifaceted analysis of how post-training reshapes a model’s competence, behavior, and execution reliability.

CapTrack decomposes LLM behavior into interpretable, user-relevant capabilities and defines an explicit mapping from tasks to capabilities and from capabilities to measurable signals. We adopt an extended notion of forgetting as any systematic drift from the base model that adversely affects capabilities, default behaviors, or execution reliability. Importantly, capability-level forgetting requires nuanced interpretation: not all changes are inherently harmful, and some reflect application-dependent trade-offs rather than pure degradation.

Conceptually, CapTrack separates latent competence from observable behavior and execution, aligning with distinctions between ability and behavior~\citep{chomsky1965aspects, firestone2020performance} and competence and performance~\citep{romeroalvarado2026capabilitiesaintneedmeasuring}. Empirically, these groups exhibit distinct drift patterns under post-training, while flat aggregation obscures these differences (App.~\ref{app:taxonomy_vs_flat}). We therefore use the taxonomy not as a universal ground truth, but as a principled and interpretable abstraction for analyzing how post-training affects distinct aspects of model behavior. 

CapTrack organizes capabilities into three complementary groups (Table~\ref{tab:captrack_eval_suite}). \textcolor{canColor}{\textbf{\textsc{CAN}}} captures latent competence: what a model can do under ideal prompting, including parametric knowledge, reasoning, contextual comprehension, faithfulness, and robustness. \textcolor{willColor}{\textbf{\textsc{WILL}}} captures default behavioral preferences: what a model tends to do by default, including willingness to answer, informational scope, verbosity, hedging, and formatting. \textcolor{howColor}{\textbf{\textsc{HOW}}} captures protocol compliance and execution: how reliably a model follows instructions and interaction protocols, including format fidelity, tool use, multi-turn consistency, long-context handling, and citation mechanics. This structure separates competence, behavior, and execution failures that would otherwise be conflated under a single aggregate notion of forgetting.

The evaluation suite operationalizes these capabilities by building on 28 established benchmarks, augmented with targeted adaptations and capability-specific metrics. These include schema-constrained output-format tests, underspecified-prompt compliance probes, rephrased robustness variants, and citation-aware QA setups, enabling measurement of behaviors not directly observable in standard benchmark settings. Large benchmarks are subsampled using Scales++~\citep{bean2025scalescomputeefficientevaluation}, yielding a suite of $\sim$19.3k samples that remains practical across model families and scales. Full details on benchmark construction, metric definitions, and judge design are provided in Appendix~\ref{app:benchmarks}--\ref{app:judge}. Metric correlations indicate low redundancy across capability dimensions (App.~\ref{app:metric_correlation}).


For \textcolor{canColor}{\textbf{\textit{latent competence} (\textsc{CAN})}}, CapTrack evaluates competence using task accuracy and targeted auxiliary signals. Parametric knowledge is measured against broad factual and skill-oriented benchmarks, while reasoning and problem-solving augment accuracy with signals such as reasoning quality and step count, which are decomposed into narrowly defined sub-criteria. In-context learning is assessed via question answering, capturing answer correctness, and evidence use. Faithfulness measures hallucination and grounding, while robustness is probed through targeted variants including prompt rephrasing, domain shift, and multilingual evaluation.

For \textcolor{willColor}{\textbf{\textit{behavioral preferences} (\textsc{WILL})}}, CapTrack measures how models respond rather than what they know. Willingness to answer is evaluated under benign, unsafe, and underspecified prompts, capturing compliance and refusal behavior. Helpfulness and informational scope are assessed using coverage and overreach metrics derived from LLM-based judgments and implemented via statement-level inclusion checks. Style and elaboration capture changes in verbosity, hedging, and formatting, reflecting shifts in default behavior rather than underlying competence.

For \textcolor{howColor}{\textbf{\textit{protocol compliance and execution} (\textsc{HOW})}}, CapTrack evaluates execution reliability using structured metrics. Instruction following is measured via constraint satisfaction, and output-format fidelity via schema adherence (parse success). Tool use is evaluated through correct selection and argument accuracy. Multi-turn commitment captures state tracking across interactions, while context-window operations assess long-context handling. Citation and attribution evaluate formatting correctness and source alignment.

We report results as relative deviations from the corresponding out-of-the-box (OOB) model, enabling comparison across heterogeneous benchmarks and metrics. This relative setup detects capability drift rather than absolute performance and reduces sensitivity to benchmark subsetting and judge calibration. Judge-based metrics cover 14 of 52 metrics; we validate their reliability through systematic comparisons across 10 judge models and additional robustness checks, finding that relative deviations are highly stable, with a mean inter-judge standard deviation of 1.1 pp, well below observed drift variance (App.~\ref{app:judge}). For interpretability, we discretize relative deviations into five regimes: \emph{None} ($>-1\%$), \emph{Minor} ($-1\%$ to $-5\%$), \emph{Moderate} ($-5\%$ to $-10\%$), \emph{Major} ($-10\%$ to $-20\%$), and \emph{Catastrophic} ($<-20\%$), while all quantitative analyses use the underlying continuous metrics.

CapTrack is designed to assess \emph{stability} rather than \emph{plasticity}. While post-training is expected to improve in-domain performance, CapTrack focuses on identifying drift in general-purpose capabilities during adaptation. In-domain results for legal and medical tasks are reported separately in Appendix~\ref{app:in-scope}. The evaluation code and dataset are available at\footnote{
\url{https://github.com/thomsonreuters/captrack}

\url{https://huggingface.co/datasets/tri-fair-lab/captrack}
}.

\section{Experiments}
\label{sec:experiments}

\subsection{Experimental Setup}
\label{subsec:experimental_setup}

We evaluate post-training effects on seven instruction-tuned LLMs from three model families: Qwen~3, Gemma~3, and Llama~3.1/3.3. The models span a broad range of scales, including industry-sized models up to 80B parameters, enabling analysis at a practical deployment scale. We group models into three size categories: \textbf{large} (70B/80B), \textbf{medium} (8–14B), and \textbf{small} (4B). Qwen~3 Next 80B follows a mixture-of-experts architecture, while all other models are dense. All experiments are conducted on instruction-tuned models, reflecting the common setting in which aligned third-party models are further adapted rather than retrained from base checkpoints (see App.~\ref{app:models}).

We study post-training in two domains, legal and medical, using two widely adopted algorithms: IFT\citep{zhang2025instructiontuninglargelanguage} and DPO \citep{rafailov2024directpreferenceoptimizationlanguage} (see App.~\ref{app:posttraining}). For each domain, we rely on curated post-training mixtures that combine general and domain-specific data. The IFT mixture contains 60k examples (50\% domain-specific), and the preference mixture contains 126k examples (40\% domain-specific). Full details on data sources and mixtures are provided in Appendix~\ref{app:data-mixtures}.

To isolate post-training effects, we use a fixed training configuration across all models and domains. Each model is trained on a HyperPod cluster with 8$\times$H200 GPU nodes for a single epoch per post-training method (for complete training settings see App. ~\ref{app:hyperparameters}). Inference is performed with vLLM~\citep{kwon2023efficient} using fixed decoding parameters (App.~\ref{app:inference}). All model variants are evaluated with CapTrack three times with different seeds to report the averaged results. 

\subsection{Forgetting Capabilities Across Post-Training Stages}

Figure~\ref{fig:forgetting_stage_comparison} provides a high-level view of forgetting at different post-training stages, aggregated by capability group and shown separately for the legal and medical domains. We report average forgetting within each capability category, with faint markers indicating benchmark-level variability.

Across both domains, forgetting extends well beyond parametric knowledge, with all capability groups exhibiting systematic drift. While latent competence (\textcolor{canColor}{\textsc{CAN}}) degrades moderately on average, the strongest effects occur in robustness-related capabilities (C5b–c) and default behavioral preferences (\textcolor{willColor}{\textsc{WILL}}), including changes in refusal behavior, response coverage, and stylistic properties. Protocol-level execution (\textcolor{howColor}{\textsc{HOW}}) is comparatively more stable, but still shows measurable degradation, particularly in multi-turn and citation scores. These results show that evaluations based on general knowledge tasks provide an incomplete picture of post-training effects, as capability-level analysis reveals behavioral and execution failures that directly impact user experience.

We further observe clear differences across post-training algorithms. IFT consistently induces the most severe forgetting across all capability groups and domains. In contrast, DPO is substantially more conservative when applied to the OOB model and can partially recover capability losses when applied after IFT, most notably within \textsc{WILL}. 
To rule out data composition as a confounding factor, we perform a controlled DPO experiment under matched training conditions. DPO retains low forgetting ($-2.71\%$ vs. $-2.43\%$) and remains substantially more stable than IFT ($-12.66\%$), indicating that these differences are not primarily driven by data composition (see App.~\ref{app:posttraining}).

We additionally evaluate GRPO on medium-scale legal-domain models (App.~\ref{app:grpo}), observing DPO-like average forgetting but more pronounced degradation in protocol-level capabilities (HOW) for some models, highlighting that post-training methods can affect capability groups differently.

\begin{tcolorbox}[  
  colback=white,  
  colframe=takeawaysColor,  
  boxrule=1.5pt,     
  arc=3pt,           
  left=4pt,right=4pt,top=4pt,bottom=4pt,  
]  
\textbf{\textcolor{takeawaysColor}{Takeaways.}}  
Post-training induces capability-level forgetting that extends well beyond parametric knowledge, with robustness and default behaviors being most affected. IFT is the primary driver of capability drift, whereas preference optimization is comparatively stable and partially recovers losses. These patterns are consistent across domains, underscoring the need for multifaceted capability-centric evaluation to understand the true impact of post-training on model behavior.  
\end{tcolorbox}

\subsection{Capability-Specific Analysis}
\begin{wrapfigure}{r}{0.55\columnwidth} 
  \centering  
  \vspace{-6pt} 
  \includegraphics[width=0.53\columnwidth]{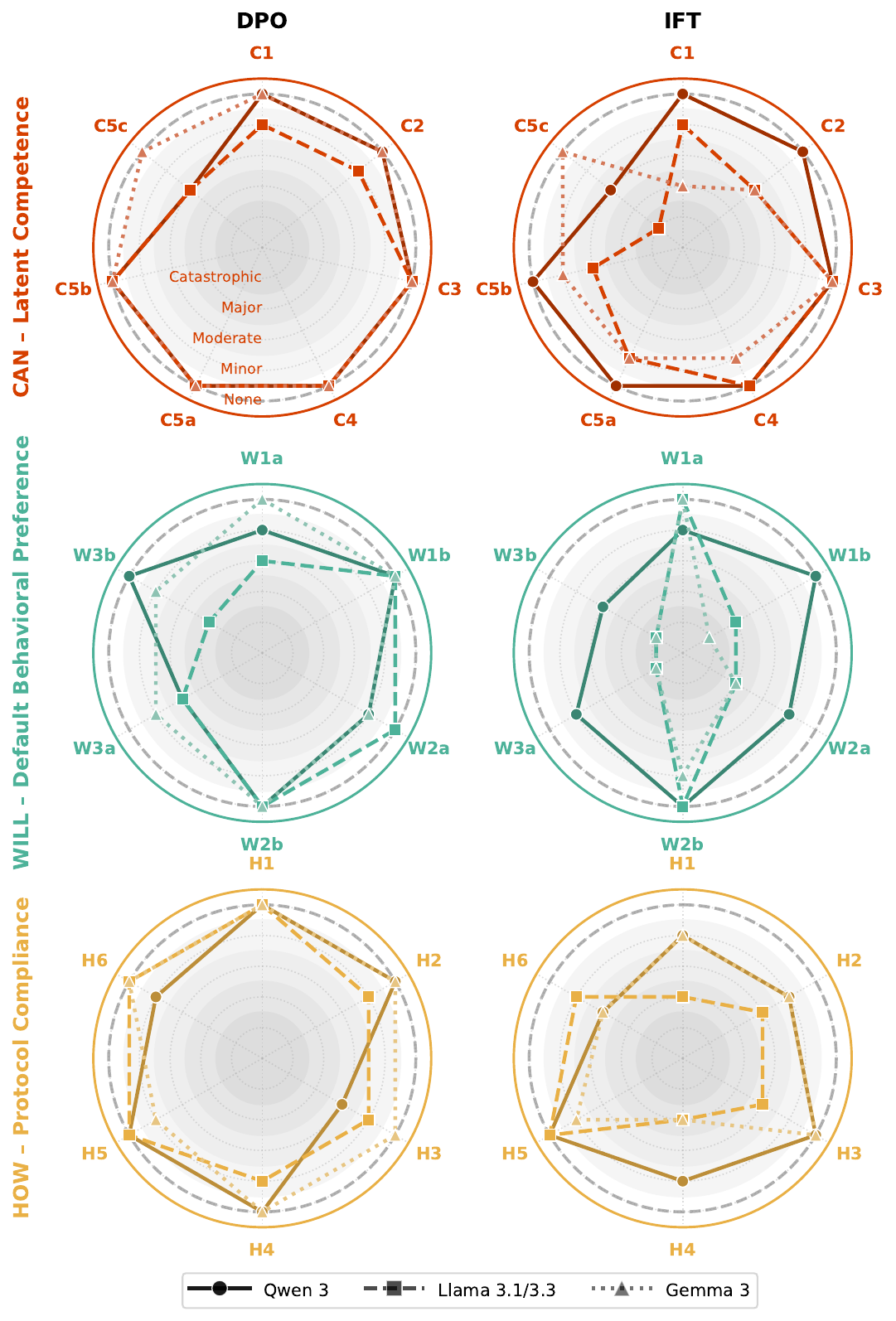}  
  \vspace{-4pt}  
  \caption{Capability-level forgetting profiles on the \textbf{legal} domain, aggregated across model sizes and shown per model family (Qwen, LLaMA, Gemma) for \textbf{IFT} and \textbf{DPO}. Further results for IFT+DPO and the medical domain are provided in Appendix~\ref{app:additional-results}. Radial distance indicates binned forgetting severity (see Section 4). Axes correspond to CapTrack sub-categories: \textsc{\textcolor{canColor}{CAN}}; C1 knowledge, C2 reasoning, C3 ICL, C4 faithfulness, C5a prompt robustness, C5b domain robustness, C5c multilingual; \textsc{\textcolor{willColor}{WILL}}; W1a unsafe refusal, W1b underspecified compliance, W2a coverage, W2b overreach, W3a verbosity, W3b formatting; \textsc{\textcolor{howColor}{HOW}}; H1 instruction following, H2 format fidelity, H3 tool use, H4 multi-turn consistency, H5 long-context, H6 citation. Results reveal strong cross-family differences, with distinct strengths and vulnerabilities across capability groups.}  
  \label{fig:main_results_spider_legal}  
  \vspace{-30pt} 
\end{wrapfigure}

We further analyze capability-level forgetting across model families, aggregating results over model sizes. Figure~\ref{fig:main_results_spider_legal} visualizes these patterns for the legal domain using the forgetting regimes from Section~\ref{sec:captrack}, with one panel per capability group and post-training algorithm and lines denoting model-family averages. This reveals systematic differences in how model families trade off stability across competence, behavior, and execution during post-training. For additional results, see Appendix~\ref{app:additional-results}.

\textbf{\textcolor{canColor}{CAN – Latent Competence.}}
Across latent competence, we observe heterogeneous but systematic forgetting. \textit{Parametric knowledge (C1)} is largely preserved under DPO, whereas IFT induces moderate to major forgetting (e.g., $-4.3\%$ for Llama, $-11.3\%$ for Gemma), driven primarily by math and code benchmarks. 
Math performance is particularly sensitive and partially recovers when DPO follows IFT, while code-related effects are less consistent. 
These degradations are concentrated in Llama and Gemma models, with Qwen remaining stable. Math-related forgetting is most pronounced in small-to-medium models, with no clear size trend for code.

A similar pattern emerges for \textit{reasoning and problem solving (C2)}: DPO causes only minor forgetting (e.g., $-1.2\%$ Llama), while IFT leads to moderate to major drops (e.g., $-8.2\%$ Llama) in reasoning score, largely driven by SuperGPQA accuracy. Beyond accuracy, IFT also degrades reasoning quality, reducing logical coherence, step consistency, and the number of reasoning steps (e.g., $-13.6\%$ for Llama). These effects again primarily affect Llama and Gemma, with Qwen largely unaffected and no consistent size dependence. In contrast, \textit{contextual comprehension (C3)} and \textit{epistemic faithfulness (C4)} show only minor forgetting across all models and settings.

Robustness-related capabilities are the most vulnerable. \textit{Domain-shift robustness (C5b)} exhibits minor to moderate forgetting under IFT (e.g., $-1.9\%$ Gemma vs.\ $-6.3\%$ Llama). \textit{Multilingual robustness (C5c)} shows the most severe degradation, with moderate forgetting under DPO (e.g., $\approx\!-9\%$) and major to catastrophic forgetting under IFT (up to $-29.4\%$ for Llama). These multilingual failures are driven mainly by Llama and Qwen models, with Gemma largely unaffected and no consistent relationship with model scale.

\textbf{\textcolor{willColor}{WILL – Default Behavioral Preferences.}}
Behavioral preferences exhibit the most pronounced and user-visible forms of forgetting. For \textit{willingness to answer under unsafe prompts (W1a)}, DPO induces minor to moderate forgetting (e.g., $-1.1\%$ Qwen, $-7.1\%$ Llama), reflecting reduced refusal of unsafe requests, primarily in Qwen and Llama models and without a clear size dependence. In contrast, \textit{compliance on underspecified tasks (W1b)} drifts far more severely: IFT causes major to catastrophic drops (e.g., $-17.5\%$ Llama, $-54.9\%$ Gemma), with models increasingly defaulting to refusal. This effect is confined to Llama and Gemma and mainly affects small and medium models.
\textit{Helpfulness and informational scope (W2)} also exhibit forgetting under post-training. After IFT, response coverage decreases across all families, ranging from minor forgetting for Qwen ($-0.9\%$) to major forgetting for Llama ($-14.1\%$) and Gemma ($-18.5\%$). DPO partially recovers coverage, most consistently for Llama ($-14.1\%\!\rightarrow\!+4.8\%$). Degradations of response coverage primarily affect medium and small models.

The strongest behavioral drift occurs in \textit{style and level of elaboration (W3)}. IFT leads to catastrophic reductions in verbosity for Gemma ($-39.3\%$) and Llama ($-35.8\%$), while Qwen shows only moderate declines ($-3.4\%$). DPO induces smaller but still consistent reductions (up to $-6.7\%$). These changes stem from shorter responses and fewer sentences, with IFT further reducing response-length variance, indicating a shift toward more uniform outputs. 
Formatting behaviors (usage rates of e.g., bullets, tables, emojis) similarly drop under IFT for Llama ($-41.3\%$) and Gemma ($-46.5\%$), with moderate effects for Qwen ($-6.9\%$), and no consistent dependence on model size.

\textbf{\textcolor{howColor}{HOW – Protocol Compliance and Execution.}}
Protocol-level capabilities are comparatively more stable than behavioral preferences, but still exhibit meaningful forgetting. 
For \textit{instruction following (H1)}, IFT induces major degradation in Llama models ($-13.5\%$) and minor losses in Qwen ($-2.1\%$) and Gemma ($-2.9\%$), which are largely recovered by DPO (e.g., $-13.5\%\!\rightarrow\!-1.7\%$ for Llama).
Lower-level constraints such as punctuation, case changes, and language switches remain largely unaffected, and no consistent size dependence is observed. 
\textit{Output-format fidelity (H2)} shows minor forgetting for Qwen and Gemma but moderate forgetting for Llama ($-9.2\%$) after IFT, primarily in small and medium models.

\textit{Tool and function use (H3)} is mostly stable, with moderate forgetting limited to Qwen after DPO ($-5.3\%$) and Llama after IFT ($-7.2\%$), driven mainly by multi-turn tool-calling failures. In contrast, \textit{multi-turn state and commitment keeping (H4)} is substantially more vulnerable: IFT causes major drops for Llama ($-16.6\%$) and Gemma ($-10.2\%$), reflecting breakdowns in state maintenance and consistency. DPO partially recovers these capabilities, though notable deficits remain. 

\textit{Context-window operations (H5)} remain largely stable across models, with only minor degradation for Gemma. Finally, \textit{citation and attribution mechanics (H6)} exhibit moderate forgetting under IFT across all families ($-4.6\%$ to $-8.2\%$), driven by reduced source accuracy, with a clear trend of increased forgetting for smaller models.

\begin{tcolorbox}[  
  colback=white,  
  colframe=takeawaysColor,  
  boxrule=1.5pt,     
  arc=3pt,           
  left=4pt,right=4pt,top=4pt,bottom=4pt,  
]  
\textbf{\textcolor{takeawaysColor}{Takeaways.}}
Capability-level forgetting varies systematically across model families, making family choice a practically relevant decision. Llama and Gemma models tend to exhibit stronger drift in latent competence and behavioral preferences, while generally more stable Qwen models are vulnerable in multilingual robustness and instruction-following. 
Larger models are not uniformly more robust to forgetting. Overall, results indicate that neither model family nor size alone consistently determines post-training stability.
\end{tcolorbox}

\section{Effectiveness of Mitigating Forgetting}
To study the drivers of capability-level forgetting and assess potential mitigation strategies, we conduct targeted ablation and intervention experiments along classical continual learning axes \citep{parisi2019continual, Delange_2021}: data, and regularization. For tractability, all experiments are restricted to the medium-sized models (see Section~\ref{subsec:experimental_setup}) and use identical post-training and evaluation settings to isolate the effect of each intervention.

\subsection{Data-Centric Mitigation: Domain-Specific vs.\ General Post-Training}
\label{sec:data_ablation}
\begin{figure*}[t]  
  \centering  
  \vspace{-6pt} 

  \begin{subfigure}[t]{0.49\textwidth}  
    \centering  
    \includegraphics[width=\linewidth]{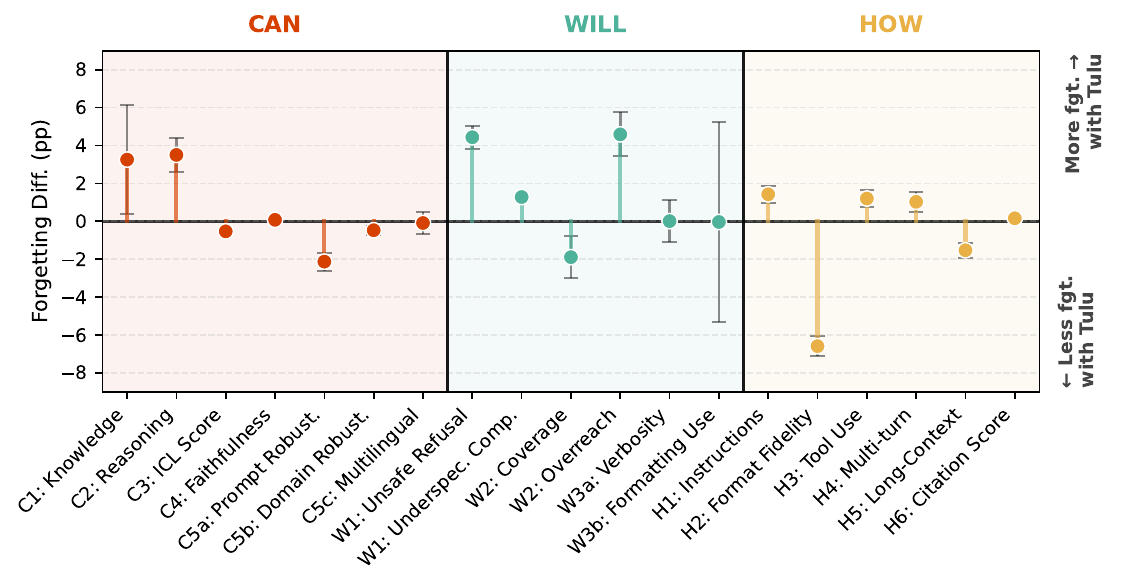}  
    \caption{Per-capability forgetting differences between general (Tulu~3) and domain-specific (legal) post-training. Points show averages across models and algorithms, with error bars showing variability. Positive values indicate less forgetting with Tulu. Effects are mixed, showing no systematic benefit of general data.}  
    \label{fig:forgetting_diff_tulu_vs_legal}  
  \end{subfigure}\hfill  
  \begin{subfigure}[t]{0.49\textwidth}  
    \centering  
    \includegraphics[width=\linewidth]{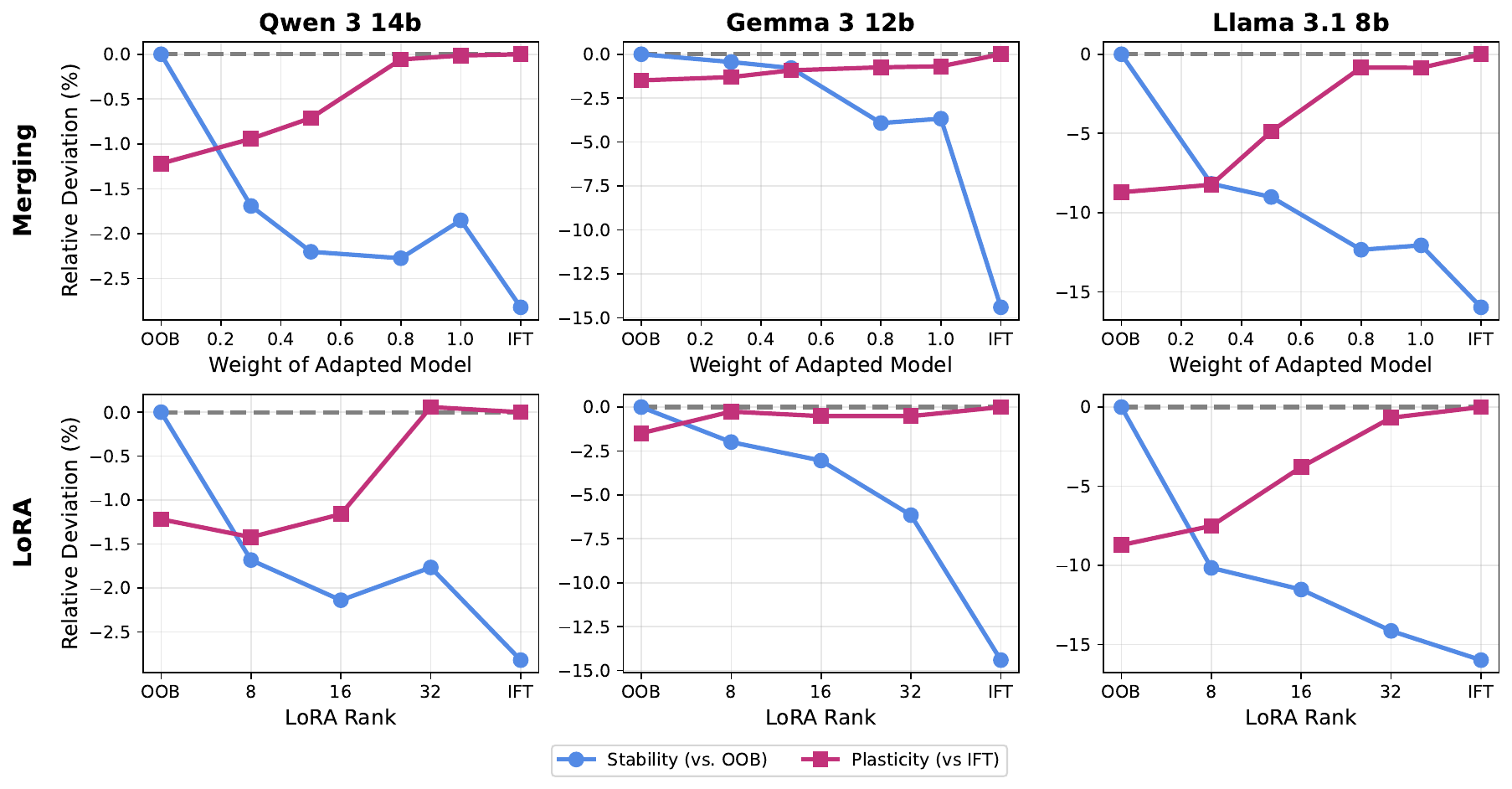}  
    \caption{Stability–plasticity trade-offs for model merging (top) and LoRA fine-tuning (bottom) on the legal domain. Stability measures average forgetting, while plasticity measures in-domain legal performance relative to unregularized IFT. Increasing regularization improves stability but consistently reduces plasticity.}   
    \label{fig:model_merging_tradeoff}  
  \end{subfigure}

  \vspace{-4pt} 
  \caption{(Left) Forgetting differences under general (Tulu~3) vs.\ domain-specific (legal) post-training. (Right) Stability--plasticity trade-offs for regularized adaptation methods on the legal domain.}  
  \label{fig:tulu_vs_legal_and_tradeoff}  
  \vspace{-12pt} 
\end{figure*}  

A natural hypothesis is that the observed forgetting arises primarily from \emph{domain-specific} post-training, rather than post-training itself. To test this, we compare forgetting induced by legal-domain post-training with forgetting induced by general, non-domain-specific post-training using Tulu~3~\citep{lambert2024tulu3} as a representative large-scale general instruction and preference dataset. We construct IFT and DPO mixtures by sampling subsets from the Tulu~3 instruction and preference data to match the size of our legal mixtures, and post-train the same models under identical settings. This isolates the effect of the data source while holding the model, algorithm, and optimization configuration fixed. Figure~\ref{fig:forgetting_diff_tulu_vs_legal} reports capability-level differences in forgetting between general and legal post-training, aggregated across model families, and post-training algorithms. Full details of mixture construction and additional ablations are provided in Appendix~\ref{app:data-ablations}.

The results show no consistent advantage of general post-training data. For latent competence (\textcolor{canColor}{\textsc{CAN}}), general data slightly increases forgetting in math and reasoning but reduces forgetting on prompt form robustness. Behavioral preferences (\textcolor{willColor}{\textsc{WILL}}) exhibit mixed effects: unsafe refusal and overreach worsen under general post-training, while coverage degrades less, and verbosity-related behaviors vary by capability. For protocol compliance (\textcolor{howColor}{\textsc{HOW}}), general post-training modestly decreases instruction following but improves output-format fidelity. These patterns closely mirror those observed in a complementary data-mixture ablation (App.~\ref{app:data-mixture-ablation}), where removing general data from the legal mixture again produces heterogeneous, capability-specific effects.

\begin{tcolorbox}[  
  colback=white,  
  colframe=takeawaysColor,  
  boxrule=1.5pt,     
  arc=3pt,           
  left=4pt,right=4pt,top=4pt,bottom=4pt,  
]  
\textbf{\textcolor{takeawaysColor}{Takeaways.}}
Data-centric interventions alone do not reliably mitigate forgetting. Capability-level forgetting is not driven by domain specificity per se, but by finer-grained properties of the post-training data. Switching to or incorporating non-domain-specific data yields mixed, capability-dependent effects and does not systematically reduce forgetting, underscoring the limits of data-only mitigation strategies.
\end{tcolorbox}
\subsection{Regularization-Based Mitigation: Model Merging \& Low-Rank Adaptation}
\label{sec:regularization}
We evaluate two mitigation strategies - model merging and parameter-efficient fine-tuning via low-rank adaptation (LoRA) \citep{hu2021loralowrankadaptationlarge, thede2025reflectingstaterehearsalfreecontinual} - both assessed in terms of plasticity (relative in-domain legal performance) and stability (average capability-level forgetting measured by CapTrack). As IFT induces the strongest forgetting, we focus on IFT-trained models throughout.
We apply model merging using TIES \citep{yadav2023tiesmergingresolvinginterferencemerging} with a sparsity density of 0.1, interpolating between a post-trained model and its corresponding OOB model. Results for alternative merging methods, hyperparameters, and additional post-training stages are provided in Appendix ~\ref{app:merging-methods}. For LoRA with a fixed learning rate, varying the LoRA rank to control the degree of parameterization. Results for different learning rates are reported in Appendix ~\ref{app:lora}.

Both methods consistently reveal the same stability–plasticity trade-off (Figure ~\ref{fig:model_merging_tradeoff}). For model merging (top row), increasing the contribution of post-trained weights improves in-domain performance but amplifies forgetting across capabilities, while stronger regularization toward the OOB model reduces forgetting at the cost of diminished plasticity. For LoRA (bottom row), higher ranks improve in-domain performance but exacerbate forgetting, while lower ranks better preserve stability at the cost of reduced adaptation. Despite constraining updates more locally than full fine-tuning, LoRA does not fundamentally change this trade-off. This behavior is consistent across model families and closely mirrors classical continual learning dynamics \citep{kirkpatrick2017overcoming}.

\begin{tcolorbox}[  
  colback=white,  
  colframe=takeawaysColor,  
  boxrule=1.5pt,     
  arc=3pt,           
  left=4pt,right=4pt,top=4pt,bottom=4pt,  
] 
\textbf{\textcolor{takeawaysColor}{Takeaways.}} Model merging and LoRA act as global regularizers that cannot decouple adaptation from forgetting, reducing capability drift at the cost of diminished in-domain gains. 
Effective mitigation of capability-level forgetting likely requires targeted, capability-aware interventions rather than stronger global constraints, highlighting a clear no-free-lunch trade-off between adaptation and retention that global regularization alone cannot resolve.

\end{tcolorbox}

\section{Conclusion}
Post-training has become the standard mechanism for adapting LLMs to new applications and domains, but is known to induce forgetting, typically studied as the loss of parametric or factual knowledge. Arguably, this view is insufficient for modern LLMs, whose usefulness depends on a broad range of interactive capabilities. Thus, we extend the notion of forgetting to encompass any systematic model drift that negatively affects latent competence, behavioral preferences, or execution reliability, and introduce \textbf{CapTrack} as a capability-centric framework for assessing such effects.

Using CapTrack, we conduct a large empirical study across multiple model families, domains, and post-training algorithms. We find that forgetting extends beyond parametric knowledge, with particularly strong drift in robustness and default behavioral preferences. Instruction fine-tuning induces substantially more forgetting than preference optimization, while DPO is comparatively conservative and can partially recover losses. Forgetting patterns vary across model families but show little consistent dependence on model size, and are not inherently domain-driven. Common mitigation strategies primarily expose a stability–plasticity trade-off rather than a universal solution.

In summary, understanding post-training effects requires moving beyond accuracy-centric evaluation toward capability-level analysis, and CapTrack provides a practical tool for exposing capability drifts.
While we apply CapTrack to study forgetting in LLM post-training, the framework is independent of the specific intervention and can equally be used to assess the effects of model compression, knowledge editing, or other model modifications. 
We discuss limitations and underlying assumptions of our framework in Appendix~\ref{sec:limitations}. 
We hope this work enables more informed model selection and adaptation and encourages future research on \emph{capability-aware} mitigation strategies targeting multifaceted forgetting.

\section*{Acknowledgements}
We thank our colleagues Nabeel Seedat, Shengzhuang Chen, and Andrew Bean for helpful discussions and feedback.
Lukas Thede thanks the International Max Planck Research School for Intelligent Systems (IMPRS-IS) for support. We are
grateful for the support of the Carl Zeiss Foundation, project ”Certification and Foundations of Safe Machine Learning Systems in Healthcare”. This work was partially funded by the ERC (853489 - DEXIM) and the Alfried Krupp von Bohlen und Halbach Foundation, which we thank for their generous support.

\section*{Impact Statement}

This work advances the evaluation of large language model post-training by introducing a capability-centric perspective on forgetting. By highlighting forms of model drift that extend beyond parametric knowledge, our findings can help practitioners better assess the reliability and behavioral stability of post-trained models in real-world settings.

CapTrack is designed as a diagnostic framework rather than a deployment mechanism. It does not introduce new modeling techniques or training procedures; instead, it provides a structured way to surface behavioral and capability changes that may otherwise remain hidden, including effects on robustness, refusal behavior, and protocol compliance. In this way, the framework may support safer and more transparent model development by improving visibility into potential failure modes.

Our experiments focus on legal and medical adaptation but do not advocate for deployment in sensitive domains. We do not release post-trained models or training data, and CapTrack measures relative capability drift rather than absolute task quality. As with any evaluation framework, individual metrics should be interpreted in context and alongside domain-specific evaluation and human oversight.

Overall, CapTrack aims to support more reliable and transparent post-training evaluation by making capability-level drift measurable and easier to analyze.

\bibliography{references_clean}
\bibliographystyle{icml2026}

\newpage
\appendix
\onecolumn

\section*{Appendix}

This appendix provides additional methodological details, analyses, and results that support the findings presented in the main paper. Its primary purpose is to improve transparency and reproducibility by documenting experimental configurations, data mixtures, evaluation procedures, and supplementary results that are referenced throughout the paper but omitted from the main text due to space constraints.

The appendix is organized into sections covering experimental setup details, post-training data mixtures, the CapTrack evaluation suite, extended experimental results, ablation studies, and mitigation experiments. Unless stated otherwise, all settings, metrics, and conventions follow those used in the main experiments. Figures and tables included here complement the main results and enable a more fine-grained inspection of capability-level forgetting and the factors that contribute to it.

\section{Experimental Setup Details}

\subsection{Models}
\label{app:models}

We evaluate a diverse set of instruction-tuned large language models drawn from three widely used model families: Qwen, Llama, and Gemma. The selected models span a broad range of parameter scales and architectural choices, enabling analysis of post-training effects across both model size and family. All models are evaluated in their publicly released instruction-tuned variants, reflecting the common practical setting in which post-training is applied to already aligned models rather than to base checkpoints. Below, we briefly describe each model and provide the corresponding HuggingFace identifiers.

\paragraph{Qwen Family.}
The Qwen models, released by Alibaba, place a strong emphasis on multilinguality, reasoning, and instruction-following capabilities \citep{qwen3technicalreport}. We include three Qwen variants:
\begin{itemize}
    \item \texttt{Qwen/Qwen3-Next-80B-A3B-Instruct} is a large-scale instruction-tuned model with approximately 80B parameters. It employs a mixture-of-experts (MoE) architecture, activating a subset of experts per token to achieve high capacity while controlling inference cost. This model represents the upper end of the scale in our study.
    \item \texttt{Qwen/Qwen3-14B} is a dense mid-sized model with approximately 14B parameters. Its standard context length is approximately 32k tokens. Following recommendations by the Qwen authors, we apply RoPE scaling at inference time to extend the effective context length to approximately 131k tokens, enabling consistent long-context evaluation across models.
    \item \texttt{Qwen/Qwen3-4B-Instruct-2507} is a small, dense instruction-tuned model with approximately 4B parameters. This model enables analysis of post-training effects at smaller scales, where capacity constraints may amplify forgetting.
\end{itemize}

\paragraph{Llama Family.}
The Llama models, released by Meta, are among the most widely adopted open-weight instruction-tuned LLMs \citep{grattafiori2024llama3herdmodels}. We include two variants:
\begin{itemize}
    \item \texttt{meta-llama/Llama-3.3-70B-Instruct} is a large-scale dense instruction-tuned model with approximately 70B parameters. It serves as a strong baseline for high-capacity models and is commonly used in both research and industrial settings.
    \item \texttt{meta-llama/Llama-3.1-8B-Instruct} is a smaller dense instruction-tuned model with approximately 8B parameters, allowing us to study post-training behavior in a lower-capacity regime within the same model family.
\end{itemize}

\paragraph{Gemma Family.}
The Gemma models, released by Google, are efficient instruction-tuned models derived from the Gemini research line \citep{gemmateam2025gemma3technicalreport}. Although Gemma 3 models are multimodal by design, incorporating both language and vision components, we use them exclusively as pure language models in this work. All multimodal inputs are disabled, and evaluations are conducted on text-only inputs. We evaluate two Gemma variants:
\begin{itemize}
    \item \texttt{google/gemma-3-12b-it} is a dense instruction-tuned model with approximately 12B parameters, representing a medium-scale model with strong instruction-following and reasoning performance.
    \item \texttt{google/gemma-3-4b-it} is a smaller dense instruction-tuned model with approximately 4B parameters, enabling comparison with similarly sized Qwen and Llama models.
\end{itemize}

Across all families, models are evaluated using their default tokenizers and context length configurations as provided by the respective HuggingFace releases, unless stated otherwise. No model-specific architectural modifications are applied prior to post-training.

\subsection{Post-Training Algorithms}
\label{app:posttraining}

We study two widely used post-training algorithms for adapting instruction-tuned language models to new domains: instruction fine-tuning (IFT) and preference optimization via direct preference optimization (DPO). Both methods are applied in a supervised post-training setting and operate on fixed datasets, without model-specific hyperparameter tuning. This subsection describes their learning objectives and training setup and clarifies the relationship between algorithmic effects and data composition.

\paragraph{Instruction Fine-Tuning.}
Instruction fine-tuning is performed using a standard supervised learning setup on instruction-response pairs. Given a dataset of prompts \(x\) and corresponding gold responses \(y = (y_1, \dots, y_T)\), the model is trained to maximize the conditional likelihood of the response given the prompt. Concretely, we minimize the negative log-likelihood (NLL) loss
\begin{equation}
\mathcal{L}_{\text{IFT}} = - \sum_{t=1}^{T} \log p_\theta(y_t \mid y_{<t}, x),
\end{equation}
where \(p_\theta\) denotes the model parameterized by \(\theta\). Training is performed with teacher forcing, and gradients are computed only with respect to the response tokens, masking the prompt portion of the sequence.

IFT directly updates the model to imitate the provided target responses and is highly effective for domain adaptation and instruction-following. However, because it treats all supervision equally and does not explicitly encode preferences or trade-offs among alternative behaviors, it can strongly reshape both latent competence and default behavioral preferences. As shown in the main paper, this often makes IFT the primary driver of capability-level forgetting.

\paragraph{Direct Preference Optimization.}
Preference optimization is performed using the standard DPO formulation \citep{rafailov2024directpreferenceoptimizationlanguage}, which optimizes a model to prefer chosen responses over rejected alternatives for a given prompt. Each training example consists of a prompt \(x\), a chosen response \(y^+\), and a rejected response \(y^-\). DPO optimizes the following objective:
\begin{equation}
\mathcal{L}_{\text{DPO}} = - \log \sigma \left( 
\beta \left[
\log \frac{p_\theta(y^+ \mid x)}{p_{\theta_0}(y^+ \mid x)}
-
\log \frac{p_\theta(y^- \mid x)}{p_{\theta_0}(y^- \mid x)}
\right]
\right),
\end{equation}
where \(\sigma(\cdot)\) denotes the logistic sigmoid, \(\beta\) controls the strength of the preference signal, and \(p_{\theta_0}\) is a fixed reference model initialized as the corresponding out-of-the-box (OOB) model. The reference model remains frozen throughout training.

Unlike reinforcement-learning–based alignment methods, DPO does not require reward modeling or on-policy sampling. Instead, it directly optimizes relative likelihoods while implicitly constraining updates through the reference model. This property makes DPO substantially more conservative than IFT and helps preserve many pre-existing capabilities, particularly at the level of default behavioral preferences.

\paragraph{Training Setup.}
For both IFT and DPO, models are trained using a fixed post-training protocol shared across all models and domains. In particular, training duration and optimization effort are held constant, with each model trained for a single pass over its respective post-training dataset and without early stopping or model-specific tuning. All hyperparameter choices are reported in Appendix~\ref{app:hyperparameters}. This design ensures that observed differences in forgetting behavior can be attributed to the post-training algorithm and data, rather than to differences in optimization intensity.

\paragraph{Algorithm-Data Entanglement.}
While IFT and DPO differ in their learning objectives, their empirical effects are inherently entangled with data composition. Instruction data directly specifies target behaviors, whereas preference data encodes relative judgments that implicitly preserve aspects of the reference model. Consequently, differences observed between IFT and DPO reflect both algorithmic properties and the structure of the available supervision.

To mitigate confounding effects, we control dataset size, training duration, and training protocol across algorithms wherever possible. Moreover, we analyze multiple data mixtures and domains and observe consistent qualitative differences between IFT and DPO across all settings. This consistency supports the interpretation that the dominant differences in forgetting behavior primarily arise from the post-training algorithm rather than from collection-specific features of individual datasets.

\paragraph{Controlled comparison of IFT and DPO.}
To isolate the effect of the post-training algorithm from confounding factors such as data composition and training scale, we perform a controlled comparison between instruction fine-tuning (IFT) and preference optimization (DPO). Specifically, we match both methods in terms of the total number of training samples, domain composition, learning rate, and effective batch size, ensuring that any observed differences are attributable to the optimization objective rather than to differences in the training setup.

\begin{figure}[t]
    \centering
    \includegraphics[width=0.8\linewidth]{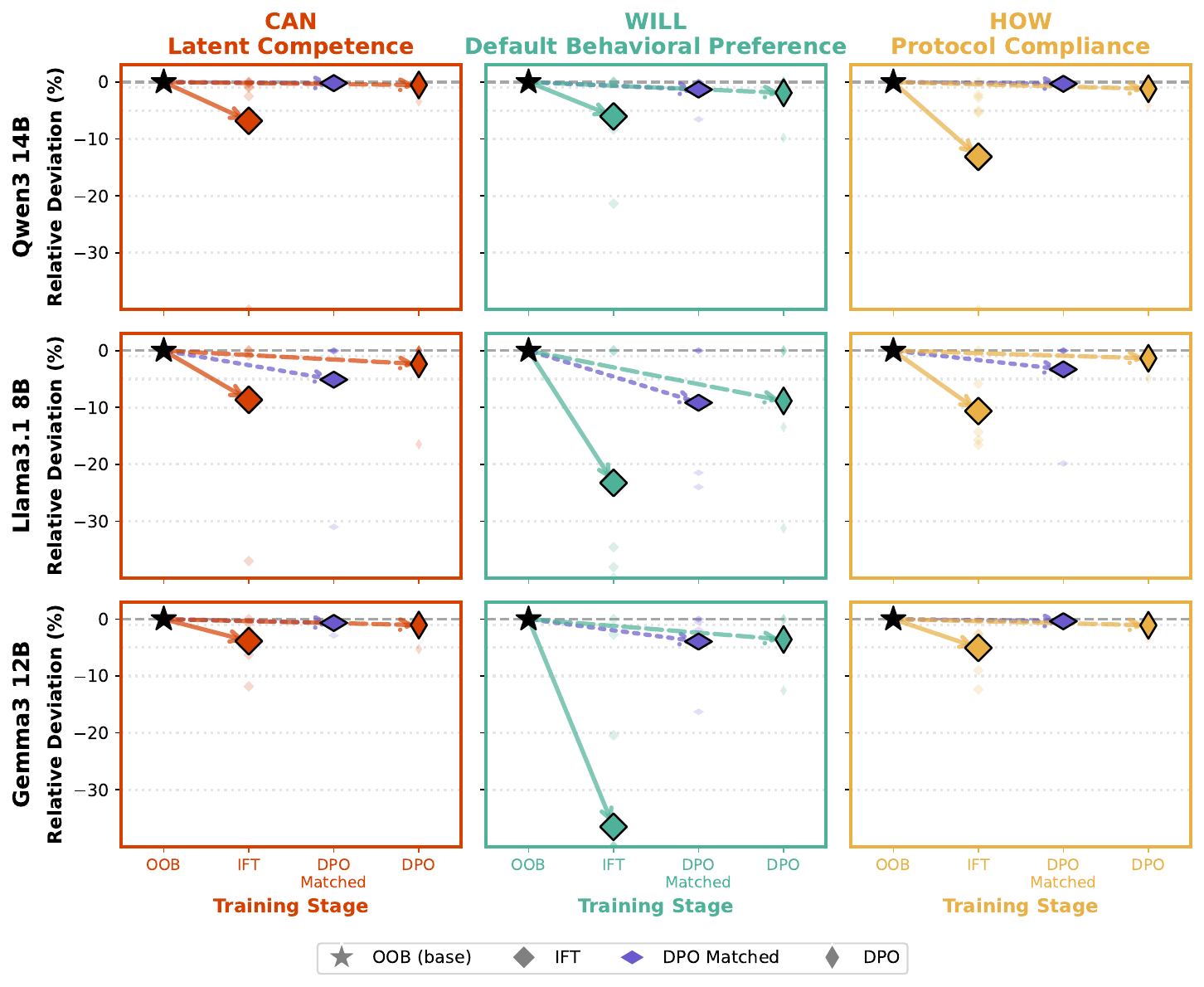}
    \caption{
    \textbf{Controlled comparison of IFT and DPO under matched training conditions.}
    Both methods are trained with identical data size, domain composition, and hyperparameters to isolate the effect of the optimization objective. 
    DPO consistently induces substantially less capability-level forgetting than IFT, confirming that the reduced drift observed in the main experiments is primarily driven by the algorithm rather than confounding factors.
    }
    \label{fig:controlled_dpo}
\end{figure}

The results confirm the trends observed in the main experiments: DPO induces substantially less capability-level forgetting than IFT, even under matched conditions. Across models, controlled DPO exhibits only minor average forgetting (e.g., $\sim$-2.5\%), closely aligning with the original DPO results, while IFT leads to significantly larger degradation (e.g., up to $\sim$-12.7\%) (Figure~\ref{fig:controlled_dpo}). These findings support the interpretation that the reduced forgetting under DPO is primarily driven by the algorithm itself, rather than differences in data or training scale.

Overall, the controlled experiments strengthen the conclusion that preference optimization acts as a more conservative post-training procedure, likely due to its implicit constraint toward a reference model, which limits capability drift during adaptation.

\subsection{Training Hyperparameters}
\label{app:hyperparameters}

\begin{table}[t]
\centering
\small
\begin{tabular}{l c c}
\toprule
\textbf{Hyperparameter} & \textbf{IFT} & \textbf{DPO} \\
\midrule
Optimizer & AdamW & AdamW \\
Learning rate & $1\times10^{-6}$ & $2\times10^{-7}$ \\
Weight decay & $0.0$ & -- \\
LR scheduler & Cosine w/ min LR & Constant w/ warmup \\
Min learning rate & $2.5\times10^{-7}$ & -- \\
Warmup steps & 10 & 10 \\
Training epochs & 1 & 1 \\
Max sequence length & 10{,}240 & 7{,}680 \\
Gradient accumulation & 8--20 & 2--7 \\
Gradient checkpointing & Enabled (non-reentrant) & Enabled (non-reentrant) \\
Max gradient norm & 1.0 & -- \\
DPO $\beta$ & -- & 0.1 \\
RPO $\alpha$ & -- & 0.2 \\
\bottomrule
\end{tabular}
\caption{
Key hyperparameters for instruction fine-tuning (IFT) and direct preference optimization (DPO).
For IFT, gradient accumulation is adjusted per model to maintain a comparable effective batch size across runs.
For DPO, the reference model is fixed to the corresponding out-of-the-box checkpoint.
}
\label{tab:post-training-hparams}
\end{table}

All post-training runs use a fixed set of hyperparameters to ensure comparability across models, domains, and post-training algorithms. We deliberately avoid model-specific tuning and train each model for a single epoch, such that observed differences in forgetting primarily reflect algorithmic and data-related effects rather than differences in optimization choices. Table~\ref{tab:post-training-hparams} summarizes the most relevant hyperparameters for IFT and DPO.

\paragraph{Common Settings.}
Across all experiments, models are trained in \texttt{bfloat16} precision using AdamW optimization with gradient checkpointing enabled. FlashAttention-2 \citep{dao2023flashattention2fasterattentionbetter} is used where supported to improve memory efficiency. All runs use a similar effective batch size, determined by the number of nodes, the per-device batch size, and the gradient accumulation rate. Random seeds are fixed across runs to ensure reproducibility.

\paragraph{Instruction Fine-Tuning.}
IFT is performed using standard next-token prediction with an NLL loss computed over response tokens only. Aside from adjustments to the gradient accumulation schedule required to maintain comparable effective batch sizes across models, all hyperparameters are held constant across domains and model families.

\paragraph{Direct Preference Optimization.}
DPO is applied using paired preference data and a frozen reference model corresponding to the out-of-the-box checkpoint. As in the IFT setting, hyperparameters are kept fixed across models and domains, and no model-specific tuning is performed.

\paragraph{Remarks.}
The relatively low learning rates reflect the post-training regime, in which models are already instruction-tuned and further adaptation should avoid large parameter updates. Learning rates differ between IFT and DPO due to their distinct optimization dynamics; rather than matching hyperparameters directly, we select learning rates such that post-trained models achieve comparable in-domain performance, enabling forgetting to be compared at similar levels of adaptation across algorithms. Ablations over learning rates in the IFT LoRA setting are provided in Appendix~\ref{app:lora-lr}. Gradient accumulation is adjusted across models solely to accommodate differences in memory footprint and sequence length. Beyond these adjustments, all hyperparameters are shared across experiments to isolate the effects of post-training algorithms and data composition.

\subsection{Inference Configuration}
\label{app:inference}

All evaluations are performed using \texttt{vLLM} as the inference backend. We adopt a fixed inference configuration across all models, domains, and post-training stages to ensure that observed performance differences reflect model changes rather than decoding or runtime effects. Unless explicitly stated otherwise, the same configuration is used for all benchmarks in the CapTrack evaluation suite.

\paragraph{General Inference Settings.}
Inference is conducted in \texttt{bfloat16} precision with tensor parallelism enabled. We configure vLLM to achieve high GPU utilization while maintaining stable execution, including for long-context inputs. Table~\ref{tab:vllm-config} summarizes the most relevant runtime parameters used throughout all evaluations.

\begin{table}[h]
\centering
\small
\begin{tabular}{l c}
\toprule
\textbf{Parameter} & \textbf{Value} \\
\midrule
Inference engine & vLLM \\
Precision & bfloat16 \\
Tensor parallelism & 8 \\
GPU memory utilization & 0.9 \\
Max model length & 130{,}000 \\
Max batched tokens & 8{,}192 \\
Max concurrent sequences & 256 \\
KV cache dtype & Auto \\
Swap space (GB) & 4 \\
\bottomrule
\end{tabular}
\caption{Key vLLM runtime configuration used for all evaluations.}
\label{tab:vllm-config}
\end{table}

\paragraph{Long-Context Configuration.}
To support long-context evaluation, we enable chunked prefill and prefix caching in vLLM, which improve throughput and memory efficiency for long prompts. For models whose default context length is insufficient for the longest benchmarks, we apply RoPE scaling at inference time. In particular, for \texttt{Qwen3-14B} we use YaRN-style RoPE scaling with a scaling factor of 4 and an original maximum position embedding length of 32k tokens, yielding an effective context length of approximately 131k tokens.

\paragraph{Decoding Parameters.}
All evaluations use deterministic decoding to minimize variance across runs. We fix the temperature to 0.0 and set \( \texttt{top-p} = 1.0 \), effectively disabling stochastic sampling. No decoding-time randomness is introduced, and all reported metrics are computed deterministically, with averaging over multiple runs applied only where required by the benchmark protocol.

\paragraph{Remarks.}
We explicitly avoid model-specific inference tuning, such as adaptive decoding strategies or per-benchmark parameter adjustments. This ensures that differences in CapTrack metrics can be attributed to post-training effects rather than inference-time optimization. The chosen configuration enables stable and scalable evaluation across both short- and long-context benchmarks, including models with up to 80B parameters.

\subsection{Compute Resources}

All post-training and evaluation experiments were conducted on an AWS HyperPod cluster using NVIDIA H200 GPUs. Post-training was performed using distributed training across 4--11 compute nodes, depending on model size and context length requirements. Large-scale evaluation and long-context inference were executed using distributed vLLM inference with tensor parallelism.

The main experimental sweep reported in the paper required approximately 47k GPU hours in total. This estimate excludes additional exploratory runs, debugging, hyperparameter tuning, and unsuccessful preliminary experiments conducted during development.

\section{Post-Training Data Mixtures}
\label{app:data-mixtures}

This appendix provides additional details on the composition and design of the post-training data mixtures used throughout our experiments. The domain-specific mixtures, particularly in legal and general settings, are not ad hoc constructions but rather carefully compiled post-training mixtures developed for real-world industrial use. Our experiments, therefore, reflect realistic post-training configurations used in practical LLM adaptation pipelines rather than synthetic or artificially curated benchmarks. We describe the legal and medical mixtures separately for IFT and DPO, followed by a description of the general post-training data shared across domains.

\subsection{Legal Domain Data Mixture}
\label{app:legal-mixture}

\paragraph{IFT Mixture.}
The legal IFT mixture combines domain-specific legal instruction data with a general replay component to stabilize general capabilities during post-training. The domain-specific portion consists of approximately 30k legal-domain examples drawn from a heterogeneous collection of legal instruction and question-answering data. These sources cover a broad range of legal tasks, including case law understanding, practical legal reasoning, legal document question answering, summarization, and applied legal analysis. Together, they span both doctrinal legal knowledge and decision-oriented legal reasoning across multiple legal subdomains.

\paragraph{DPO Mixture.}
The legal DPO mixture is designed to align model behavior with desirable legal response characteristics, such as correctness, clarity, and appropriateness. The domain-specific component comprises approximately 51k legal-preference examples derived from paired comparisons of legal-domain responses. These preferences cover tasks such as case law analysis, legal document question answering, practical legal reasoning, and summarization, as well as explicitly annotated legal preference data.

\subsection{Medical Domain Data Mixture}
\label{app:medical-mixture}

\paragraph{IFT Mixture.}
The medical IFT mixture combines domain-specific medical instruction data with a general replay component to stabilize general capabilities during post-training. The domain-specific component is primarily drawn from the \emph{Medical Meadow} collection \citep{han2025medalpacaopensourcecollection}, which aggregates a diverse set of medical instruction-style datasets covering medical knowledge acquisition, question answering, and clinical reasoning. These include medical curriculum flashcards, health advice classification, open-domain medical question answering (with and without additional context), and medical causality detection. Together, these sources span both factual medical knowledge and decision-oriented medical reasoning tasks. From the available corpora, we subsample and filter instances to obtain a balanced medical mixture of approximately 30k examples.

\paragraph{DPO Mixture.}
The medical DPO mixture is designed to align model behavior with desirable medical response characteristics, with a particular focus on advice quality and safety. The domain-specific component is drawn from a large-scale medical preference dataset based on \emph{AskDocs}-style interactions \citep{li2025aligningllmsaskgood}, where candidate responses are paired and annotated according to their suitability for providing medical advice. Preference annotations reflect multiple criteria, including medical correctness, clarity and organization, empathy and tone, and overall safety and appropriateness. From the available corpus, we subsample and reweight the data using fixed scaling factors to obtain an effective dataset size of approximately 50k preference pairs.

\subsection{General Post-Training Data}
\label{app:general-mixture}

\paragraph{General IFT Mixture.}
The general IFT mixture is shared across all domains and is used as replay data during domain-specific post-training. It comprises instruction-following examples covering general knowledge, coding, reasoning, news comprehension, multi-hop question answering, and general chat-style interactions. These data points are drawn from widely used public instruction datasets and are included to preserve the broad capabilities learned during pretraining. For domain-specific IFT experiments, this mixture contributes approximately 30k examples.

\paragraph{General DPO Mixture.}
The general DPO mixture is likewise shared across domains and consists of preference-style supervision covering general chat behavior, reasoning, and coding. It includes both single-turn preference data and reasoning-oriented supervision, such as chain-of-thought-style annotations. For domain-specific DPO experiments, approximately 75k general preference examples are included and balanced via per-dataset scaling factors to prevent the general signal from overwhelming domain-specific preferences.

\section{CapTrack Evaluation Suite Details}
\label{app:captrack}

This appendix provides additional details on the CapTrack evaluation suite, complementing the overview presented in Section \ref{sec:captrack} of the main paper. We describe the capability taxonomy, benchmark adaptations, metric definitions, and evaluation protocols in greater depth to clarify how individual capabilities are operationalized and measured.

\subsection{Capability Taxonomy (Extended)}
\label{app:taxonomy}

CapTrack organizes model behavior into three complementary capability groups (\textsc{CAN}, \textsc{WILL}, and \textsc{HOW})that together provide a structured and interpretable abstraction of how LLMs behave before and after post-training. The taxonomy is not intended to be exhaustive or uniquely correct; rather, it reflects a pragmatic design choice aimed at separating conceptually distinct sources of model drift that are difficult to disentangle through accuracy-centric evaluation alone. In particular, it disentangles latent competence from default behavioral tendencies and execution reliability, enabling fine-grained analysis of capability-level changes that would otherwise be obscured by aggregate metrics.

While alternative decompositions of LLM capabilities are certainly possible, we found this grouping to strike a useful balance between interpretability, coverage of user-relevant behavior, and empirical measurability across a wide range of models and post-training settings.

\textbf{\textcolor{canColor}{CAN -- Latent Competence:}}

The \textsc{CAN} group captures what a model is capable of doing under ideal prompting conditions, abstracting away from its default policies or stylistic preferences. These capabilities align most closely with traditional notions of model performance and generalization.

\textbf{Parametric Knowledge \& Skills (C1)} measures the preservation of general world knowledge and core skills acquired during pretraining, including mathematics, coding primitives, and elements of formal or symbolic reasoning. Degradation in this category reflects interference with the model’s internalized representations rather than surface-level behavioral changes.

\textbf{Reasoning \& Problem Solving (C2)} assesses the reliability of multi-step inference, including decomposition into subproblems, correctness of intermediate steps, arithmetic and symbolic manipulation, and causal or logical consistency. This capability captures whether a model can still reason coherently even when final answers remain superficially plausible.

\textbf{Contextual Comprehension (In-Context Learning Capacity) (C3)} evaluates the model’s ability to use information provided directly in the prompt. This includes identifying relevant evidence spans, combining multiple pieces of context, and drawing correct conclusions based on information contained in the prompt rather than relying on parametric recall.

\textbf{Epistemic Faithfulness \& Grounding (C4)} focuses on whether model outputs remain tied to available evidence and avoid unsupported additions or hallucinations. Importantly, this capability is defined independently of explicit citation or formatting requirements and instead targets the factual grounding of claims.

\textbf{Robustness of Competence (C5)} captures the stability of latent competence under benign perturbations. We decompose robustness into three sub-dimensions: \emph{C5a. Prompt-Form Invariance}, which measures sensitivity to paraphrasing or changes in prompt order and format; \emph{C5b. Domain-Shift Robustness}, which probes performance on adjacent topics or benign out-of-distribution inputs within the model’s pretraining scope; and \emph{C5c. Multilingual and Code-Switching Stability}, which evaluates whether previously acquired cross-lingual abilities are retained after post-training.

\textbf{\textcolor{willColor}{WILL -- Default Behavioral Preferences:}}

The \textsc{WILL} group characterizes what a model tends to do by default when generating responses, reflecting policy-level and alignment-induced behaviors rather than underlying competence. Changes in this group often correspond to shifts in user experience, even when latent abilities remain intact.

\textbf{Willingness to Answer (W1)} measures whether the model engages with a task at all and how it refuses when it does not. This includes both over-refusal in benign settings and under-refusal in unsafe or underspecified contexts.

\textbf{Helpfulness \& Informational Scope (W2)} captures the breadth and completeness of information the model defaults to providing. Degradation in this capability may manifest as overly concise, incomplete, or evasive responses, while improvements may increase coverage at the risk of overreach.

\textbf{Style \& Level of Elaboration (W3)} describes the model’s default presentation style, including verbosity, conciseness, hedging behavior, tone, and the use of formatting elements such as bullet points or structured layouts. Although stylistic changes may be desirable in isolation, they often correlate with broader shifts in behavior induced by post-training.

\textbf{\textcolor{howColor}{HOW -- Protocol Compliance and Execution:}}

The \textsc{HOW} group captures how reliably a model executes instructions and adheres to interaction protocols, independent of whether it possesses the underlying competence to solve a task.

\textbf{Instruction Following \& Constraint Satisfaction (H1)} evaluates the model’s ability to respect explicit rules, step-by-step instructions, and constraints such as length limits or formatting requirements.

\textbf{Output-Format Fidelity (H2)} measures whether the model produces strictly valid structured outputs (e.g., JSON objects or tables) and adheres to specified schemas or contracts.

\textbf{Tool/Function Use \& Integration (H3)} assesses the model’s ability to select appropriate tools, issue correct calls with valid arguments, and correctly integrate tool outputs into its final response.

\textbf{Multi-Turn State \& Commitment Keeping (H4)} captures whether the model maintains goals, constraints, and commitments across dialogue turns, including honoring prior instructions or decisions made earlier in the interaction.

\textbf{Context-Window Operations (H5)} evaluates the mechanics of operating over long inputs, including accurate referencing, quote usage, pointer-chasing, and summarization without loss of critical details.

\textbf{Citation/Attribution Mechanics (H6)} measures whether, when explicitly asked to cite sources, the model produces correctly formatted references and links claims to appropriate sources. This capability is distinct from epistemic faithfulness, which concerns grounding independent of citation format.

Taken together, this taxonomy supports a nuanced interpretation of post-training effects by separating changes in competence, behavior, and execution. Rather than claiming completeness, CapTrack provides a structured lens that makes different forms of model drift explicit and comparable, enabling targeted, capability-aware analysis of forgetting.

\subsection{Benchmark Modifications}
\label{app:benchmarks}

CapTrack combines established benchmarks with targeted modifications to probe specific capabilities in a controlled and comparable manner. All benchmarks are evaluated relative to the corresponding out-of-the-box (OOB) model, and large benchmarks are subsampled (via Scale++ \citep{bean2025scalescomputeefficientevaluation}) to ensure computational feasibility while preserving diversity and difficulty. Table~\ref{tab:captrack-benchmarks} provides a complete overview of all benchmarks used in the evaluation suite, including the capabilities they target, metrics, dataset sizes, evaluation subset sizes, and HuggingFace dataset identifiers.

\begin{table}[h]
\centering
\small
\resizebox{\textwidth}{!}{
\begin{tabular}{l l l r r l l}
\toprule
\textbf{Benchmark} & \textbf{Capability} & \textbf{Metric} & \textbf{\#Samples} & \textbf{Subset} & \textbf{HF Path} & \textbf{Citation} \\
\midrule
MMLU-Pro & C1, C5, H2 & Acc., schema pass & 12{,}000 & 1{,}000 & TIGER-Lab/MMLU-Pro & \citep{wang2024mmluprorobustchallengingmultitask} \\
PopQA & C1 & Acc. & 14{,}300 & 1{,}000 & akariasai/PopQA & \citep{mallen2023llm_memorization} \\
GSM8K & C1, C5, H2 & Acc., schema pass & 1{,}320 & 500 & openai/gsm8k & \citep{cobbe2021gsm8k} \\
LiveMathBench & C1 & Acc. & 100 & 100 & opencompass/LiveMathBench & \citep{liu2024your} \\
HumanEval & C1 & Acc. & 164 & 164 & openai/openai\_humaneval & \citep{chen2021evaluating} \\
MBPP & C1 & Acc. & 427 & 427 & Muennighoff/mbpp & \citep{austin2021program} \\
MATH & C2 & Acc., reason.\,$^\dagger$, steps & 500 & 500 & nlile/hendrycks-MATH-benchmark & \citep{hendrycks2021measuringmathematicalproblemsolving} \\
SuperGPQA & C2 & Acc., reason.\,$^\dagger$, steps & 26{,}500 & 500 & m-a-p/SuperGPQA & \citep{pteam2025supergpqascalingllmevaluation} \\
HotpotQA & C3, H6 & Acc., evidence hit$^\dagger$, cite pass, src. acc. & 7{,}410 & 800 & hotpotqa/hotpot\_qa & \citep{yang2018hotpotqa} \\
BoolQ & C3 & Acc., evidence hit$^\dagger$ & 3{,}270 & 500 & google/boolq & \citep{clark2019boolq} \\
RAGTruth & C4, W2 & Halluc.\,$^\dagger$, coverage$^\dagger$, overreach$^\dagger$ & 2{,}700 & 500 & wandb/RAGTruth-processed & \citep{niu-etal-2024-ragtruth} \\
TruthfulQA & C4 & Acc. & 817 & 817 & domenicrosati/TruthfulQA & \citep{lin2022truthfulqameasuringmodelsmimic} \\
WinoGrande & C5 & Acc. & 1{,}770 & 1{,}000 & allenai/winogrande & \citep{ai2:winogrande} \\
HellaSwag & C5 & Acc. & 10{,}000 & 1{,}000 & Rowan/hellaswag & \citep{zellers2019hellaswag} \\
MGSM & C5 & Acc. & 2{,}750 & 500 & juletxara/mgsm & \citep{shi2022language} \\
XTREME & C5c & Acc. & 3{,}000 & 500 & google/xtreme & \citep{hu2020xtreme} \\
HarmBench & W1 & Benign ref.$^\dagger$, unsafe comp.$^\dagger$ & 200 & 200 & walledai/HarmBench & \citep{mazeika2024harmbench} \\
ELI5 & W2 & Coverage$^\dagger$, overreach$^\dagger$ & 325{,}000 & 500 & sentence-transformers/eli5 & \citep{fan2019eli5} \\
MT-Bench & W3, H4 & Verb., hedging, direct., MT follow$^\dagger$ & 80 & 80 & philschmid/mt-bench & \citep{zheng2023judging} \\
OASST1 & W3 & Verb., hedging, direct., format & 188 & 188 & OpenAssistant/oasst1 & \citep{köpf2023openassistantconversationsdemocratizing} \\
IFEval & H1 & Pass rate & 541 & 541 & google/IFEval & \citep{zhou2023instructionfollowingevaluationlargelanguage} \\
FollowBench & H1 & Pass rate$^\dagger$ & 1{,}850 & 500 & YuxinJiang/FollowBench & \citep{jiang2023followbench} \\
BFCL & H3 & Select. acc., arg. acc. & 480 & 480 & gorilla-llm/BFCL & \citep{berkeley-function-calling-leaderboard} \\
MNMS & H3 & Select. acc., arg. acc. & 882 & 882 & zixianma/mnms & \citep{ma2024mms} \\
StructFlowBench & H4 & MT follow$^\dagger$ & 155 & 155 & Jinnan/StructFlowBench & \citep{li2025structflowbench} \\
RULER & W1, H5 & Compliance, acc. & 500 & 500 & self-long/RULER-llama3-1M & \citep{hsieh2024ruler} \\
LongBench-V2 & H5 & Acc. & 305 & 305 & zai-org/LongBench-v2 & \citep{bai2024longbench2} \\
QASPER & H6 & Cite pass, src. acc. & 402 & 402 & allenai/qasper & \citep{Dasigi2021ADO} \\
\midrule
& & & \textbf{Total:} & 14,541 & & \\
\bottomrule
\end{tabular}
}
\caption{
Complete overview of benchmarks used in the CapTrack evaluation suite.
Capabilities follow the taxonomy in Appendix~C.1.
Metrics marked with $\dagger$ are computed using LLM-as-a-judge evaluation.
``Subset'' denotes the number of samples used for evaluation.
}
\label{tab:captrack-benchmarks}
\end{table}

\paragraph{Benchmark Subsetting and Filtering.}
For several benchmarks, we apply additional filtering to ensure consistency across models and feasibility at long context lengths. For LongBench-V2, we restrict evaluation to samples with total context length below 125k tokens, enabling stable inference under the chosen long-context configuration. For MGSM, we evaluate a multilingual subset consisting of the following languages: \{\texttt{bn}, \texttt{de}, \texttt{es}, \texttt{fr}, \texttt{ja}, \texttt{ru}, \texttt{sw}, \texttt{te}, \texttt{th}, \texttt{zh}\}. For XTREME, we use the MLQA setting with the language pairs \{\texttt{MLQA.ar.ar}, \texttt{MLQA.de.de}, \texttt{MLQA.es.es}, \texttt{MLQA.hi.hi}, \texttt{MLQA.vi.vi}, \texttt{MLQA.zh.zh}\}. For OASST1, we restrict evaluation to English conversations and use only the first user prompt per conversation to avoid multi-turn confounding.

\paragraph{Prompt Reformulation and Robustness Construction.}
To probe prompt-form invariance and robustness-related capabilities, we construct rephrased variants of selected benchmarks (e.g., MMLU-Pro and GSM8K). Rewritten prompts are generated by prompting \texttt{llama-maverick} to produce semantically equivalent but syntactically distinct versions of the original prompts, while preserving task intent and required output constraints. The same reformulated prompts are used consistently across all models to ensure fair comparison.

To evaluate \emph{output-format fidelity}, we additionally wrap samples from GSM8K and MMLU-Pro in two distinct JSON schemas, requiring models to produce strictly valid structured outputs that conform to the specified schema.

To assess \emph{willingness to answer under underspecification}, we adapt the RULER-4k benchmark. RULER-4k is originally a needle-in-a-haystack task in which the target question appears at the end of the prompt. We construct underspecified variants by truncating the final 100 tokens of each prompt, thereby removing the explicit question while preserving the surrounding context. In this setting, compliant behavior requires recognizing the task's underspecification and responding appropriately, rather than hallucinating an answer or refusing to answer.

These benchmark modifications allow CapTrack to isolate specific capability dimensions, such as robustness, behavioral preferences, and protocol compliance, while maintaining comparability with prior work and controlling evaluation cost.

\paragraph{Licensing and Terms of Use.}
CapTrack combines and reformats samples from existing evaluation benchmarks. All original datasets, benchmarks, and models are credited throughout the paper and retain their respective licenses and terms of use. The released CapTrack benchmark is distributed under the Apache 2.0 license for the dataset structure, formatting, and CapTrack-specific modifications, while individual benchmark subsets remain subject to their original licenses and usage restrictions. CapTrack redistributes only samples permitted under the corresponding source licenses and is intended strictly for research and evaluation purposes.

\subsection{Metric Definitions}
\label{app:metrics}

This section defines the metrics used in the CapTrack evaluation suite. Metrics are grouped by capability category and metric family, as many benchmarks share common evaluation logic. Where applicable, we distinguish between rule-based metrics and metrics computed using an LLM-as-a-judge. 

Across all metrics computed using an LLM-as-a-judge, CapTrack follows a strict design principle of minimizing interpretive freedom. Judge-based evaluations are decomposed into narrowly scoped, atomic decisions (e.g., per-step validity, per-claim support) rather than holistic response-level judgments. Judges are prompted to assess a single, well-defined criterion at a time, and final metric values are computed deterministically by aggregating these atomic decisions. During metric construction, we perform manual spot checks to verify prompt correctness, failure modes, and consistency across benchmarks.

\subsubsection{Accuracy-Based Metrics}
Accuracy-based metrics measure task-level correctness under explicit answer formats and are computed using deterministic, rule-based procedures.

\paragraph{Multiple-Choice Accuracy.}
For benchmarks such as MMLU-Pro, SuperGPQA, TruthfulQA, WinoGrande, HellaSwag, and LongBench-V2, model outputs are evaluated by extracting a single option identifier (e.g., A/B/C/D). An example is counted as correct if the extracted option exactly matches the gold label after normalization. If the output does not conform to the expected format, correctness is determined using an LLM judge that compares the model output to the gold answer. In these cases, the judge is restricted to equivalence checking between the model output and the gold answer, rather than subjective correctness assessment.

\paragraph{Numeric Answer Accuracy.}
For math benchmarks including GSM8K, MGSM, and LiveMathBench, numeric answers are extracted from the model output using a prioritized set of patterns (e.g., explicit ``Answer:'' lines or final numeric expressions). An example is counted as correct if the extracted value matches the gold answer within a small tolerance. For problems with symbolic or non-trivial expressions, evaluation falls back to equivalence checking via symbolic normalization or an LLM judge. Here, the judge is used exclusively for symbolic equivalence checking when rule-based normalization is insufficient.

\paragraph{Code Execution Accuracy.}
For coding benchmarks (HumanEval, MBPP), generated code is executed in a clean environment against the provided unit tests. An example is considered correct if the code executes without errors and passes all tests.

\subsubsection{Reasoning Quality Metrics}
Reasoning quality metrics evaluate the structure and consistency of intermediate reasoning, independently of final-answer correctness. These metrics are computed using an LLM-as-a-judge and reported on a continuous scale in $[0,1]$. Each component is evaluated independently using dedicated prompts that constrain the judge to the specified criterion, thereby preventing a sample-level assessment of reasoning quality.

\paragraph{Average Reasoning Score.}
For MATH and SuperGPQA, the average reasoning score aggregates three per-item components: (i) step validity, measuring whether individual reasoning steps contain meaningful and valid reasoning operations; (ii) logical coherence, measuring whether consecutive steps follow logically from one another; and (iii) intermediate consistency, measuring whether intermediate values and claims remain consistent and align with the final answer. Scores are averaged per item and then across the dataset.

\paragraph{Step Statistics.}
We additionally report the average number of parsed reasoning steps per item as a descriptive statistic. This metric is not used for evaluation and is intended to contextualize changes in reasoning verbosity or decomposition depth.

\subsubsection{Evidence Use and Grounding Metrics}
These metrics assess whether model outputs remain grounded in the provided evidence and avoid unsupported claims.

\paragraph{Evidence Hit Rate.}
For contextual comprehension benchmarks (HotpotQA, BoolQ), evidence hit rate measures whether the model’s final answer is supported by the provided context. Support is determined via string-level overlap, entity matching, or approximate lexical coverage. The metric is computed as the fraction of examples with a positive evidence hit.

\paragraph{Hallucination and Overreach Rates.}
For RAGTruth and ELI5, the hallucination and overreach metrics quantify the proportion of model output claims that are not supported by the provided context (hallucination) or by the gold answer (overreach). Claims are extracted automatically and evaluated individually using an LLM judge. To maintain consistent metric directionality, with higher values indicating better performance, we report \emph{faithfulness} as $1 - \text{hallucination}$ and the \emph{overreach score} as $1 - \text{overreach}$. Judgment is performed on a single extracted claim at a time, and judges never see the full response when scoring individual claims.

\paragraph{Coverage Score.}
The coverage score measures how completely the model's output covers the key claims in the gold answer. Atomic claims are extracted from the gold answer and checked for support in the model output using an LLM judge. The metric is averaged across examples and lies in $[0,1]$. 

\subsubsection{Behavioral and Preference Metrics (WILL)}
Behavioral metrics characterize the model’s default response tendencies rather than task-level competence.

\paragraph{Refusal and Compliance Rates.}
For GSM8k, HarmBench, and RULER (4k cropped), refusal and compliance behaviors are classified using an LLM judge. The benign refusal rate measures how often the model refuses safe requests, whereas the unsafe compliance rate measures how often it complies with unsafe requests. The underspecified compliance rate measures how often the model attempts to respond to the underspecified task rather than refusing to answer. Classification prompts are formulated as closed-set decisions (e.g., refuse vs.\ attempt to answer), avoiding open-ended preference judgments.

\paragraph{Style and Elaboration Metrics.}
For MT-Bench and OASST1, we report descriptive statistics capturing default response style, including average answer length, average sentence length, and total number of sentences. Hedging rate measures the proportion of responses that contain uncertainty markers or disclaimers. Directness measures whether the first sentence of the response provides a direct answer. Formatting usage rates track the presence of bullets, tables, and emojis.

\subsubsection{Protocol and Execution Metrics (HOW)}
Protocol metrics evaluate whether the model correctly executes instructions and adheres to formal constraints.

\paragraph{Instruction Following.}
For IFEval and FollowBench, instruction-following accuracy is computed as the fraction of examples for which all required constraints are satisfied. Partial compliance does not constitute success. For IFEval, we use the original rule-based evaluation protocol and employ an LLM-as-a-judge to assess instruction-following on FollowBench.  

\paragraph{Output-Format Fidelity.}
For schema-wrapped variants of MMLU-Pro and GSM8K, format fidelity is measured as the parse or schema pass rate, i.e., whether the model output can be parsed as valid JSON or table output and satisfies all required fields.

\paragraph{Tool and Multi-Turn Execution.}
For BFCL and MNMS, tool-use performance is measured by accuracy in both tool selection and argument specification. For multi-turn benchmarks, including MT-Bench and StructFlowBench, we assess a model’s ability to maintain instructions, constraints, and commitments across dialogue turns. Concretely, an LLM judge evaluates the model’s response to a follow-up query in the context of the preceding interaction, scoring whether the second-turn answer remains consistent with earlier instructions and appropriately integrates prior context. This evaluation captures failures such as dropped constraints, inconsistent commitments, or loss of task context across turns.

\paragraph{Long-Context Operations and Citation Mechanics.}
For RULER and LongBench-V2, both of which are formulated as question-answering tasks over long input contexts, we measure accuracy by comparing the model’s final answer against the gold response. For QASPER and HotpotQA, citation metrics assess whether citations are correctly formatted and point to sources that substantiate the answer. These metrics assess the mechanical correctness of citations and are distinct from epistemic grounding, which is captured by CAN-level metrics.

\subsubsection{LLM-as-a-Judge Protocol}
Metrics marked as judge-based use a fixed LLM judge with standardized prompts across benchmarks. The judge never receives information about the evaluated model or its training condition. Judge outputs are binary labels or scalar scores in $[0,1]$, aggregated at the example and dataset levels. We use the same judge configuration across all experiments to ensure consistency and comparability. Judge prompts are designed to be strictly task-local: the judge is provided only with the minimal information required for the specific decision (e.g., a single reasoning step or factual claim), and never with benchmark-level context or aggregate scores. The judge is instructed to ignore stylistic preferences and to base decisions solely on the explicitly stated criterion. Final metric values are never produced directly by the judge, but are instead computed by aggregating atomic judge outputs using fixed, deterministic rules. This design reduces variance arising from prompt interpretation and judge preferences.

\subsection{Taxonomy-Based vs. Flat Aggregation Analysis}
\label{app:taxonomy_vs_flat}

A common approach to evaluating post-training effects is to aggregate performance across tasks into a single scalar metric. While convenient, such aggregation implicitly assumes that all benchmarks measure a homogeneous notion of model quality. In contrast, CapTrack decomposes evaluation into capability groups, enabling analysis of heterogeneous effects across different aspects of model behavior.

\paragraph{Empirical comparison.}
To quantify the impact of aggregation, we compare capability-level results with their corresponding flat averages across all benchmarks. We observe that flat aggregation systematically masks structured variation in post-training effects. In particular, models that exhibit near-zero average drift often show substantial degradation in specific capability groups, most notably in robustness, behavioral preferences, and multi-turn consistency.

\begin{figure}[t]
    \centering
    \includegraphics[width=0.9\linewidth]{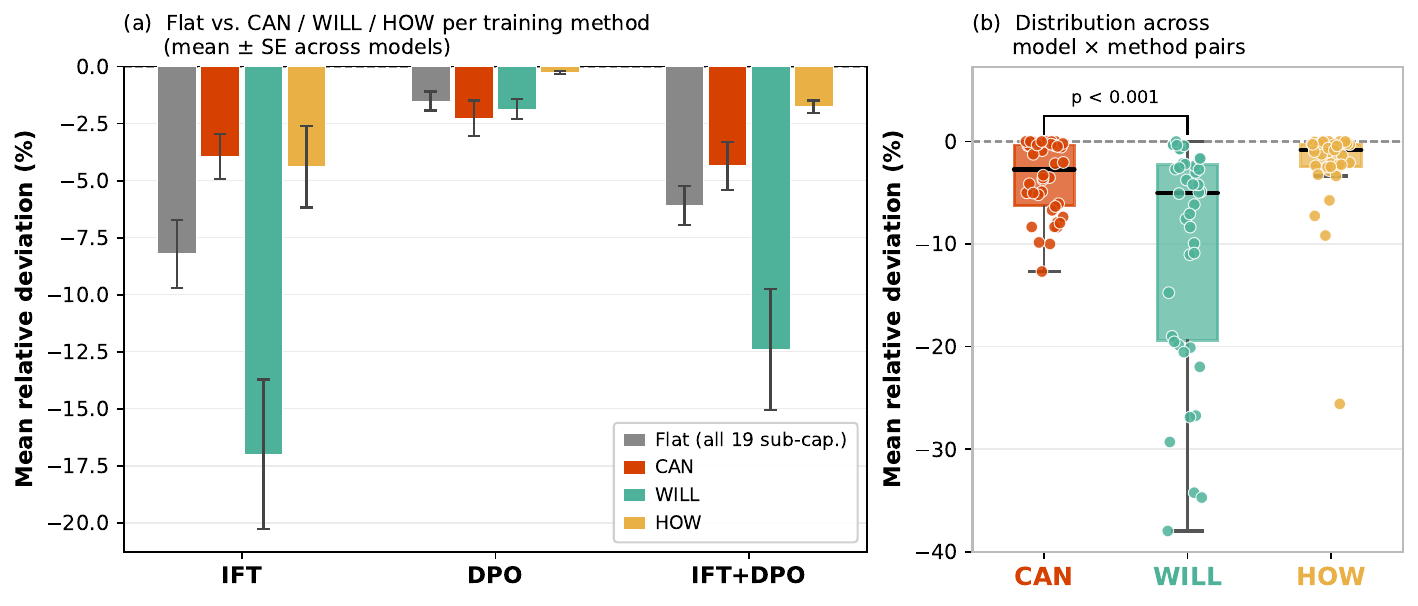}
    \caption{
    \textbf{Flat aggregation vs. capability-level analysis.}
    Comparison of average performance across all benchmarks (flat aggregation) with capability-group-specific deviations. 
    While flat averages suggest minimal overall drift, capability-level analysis reveals substantial degradation in specific dimensions, particularly in behavioral and protocol-level capabilities.
    }
    \label{fig:taxonomy_vs_flat}
\end{figure}

\paragraph{Statistical analysis.}
To formally assess whether these differences are statistically significant, we perform a paired Wilcoxon signed-rank test comparing per-model flat averages against capability-group-specific deviations. Across models and post-training settings, we find that capability-level deviations differ significantly from flat aggregates (p $<$ 0.01), indicating that aggregation obscures meaningful structure in the data.

\paragraph{Implications.}
These results demonstrate that post-training-induced forgetting is not well captured by a single scalar metric. Instead, it manifests heterogeneously across distinct capability groups, motivating the need for structured evaluation frameworks such as CapTrack. By explicitly separating latent competence, behavioral preferences, and execution reliability, CapTrack enables detection of trade-offs that remain invisible under flat aggregation.

\subsection{Metric Correlation Analysis}
\label{app:metric_correlation}

\begin{figure}[t]
    \centering
    \includegraphics[width=0.9\linewidth]{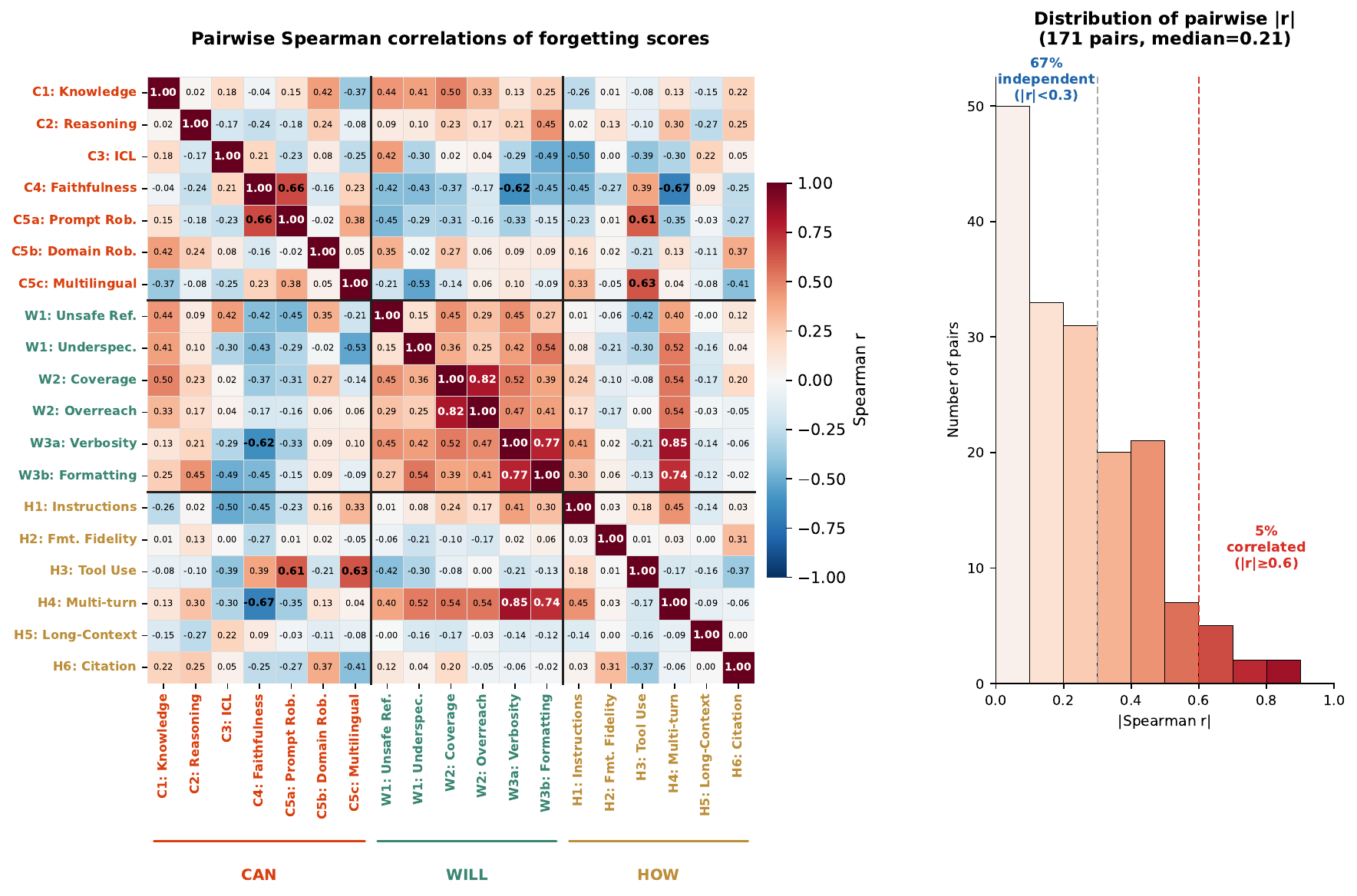}
    \caption{
    \textbf{Correlation between CapTrack sub-capabilities.}
    Pairwise Spearman correlations are computed over relative deviations across model, domain, and post-training settings. 
    Most capability pairs show weak correlations, indicating low redundancy across the evaluation suite. 
    Stronger correlations form interpretable clusters, such as verbosity, formatting use, and multi-turn commitment, or faithfulness and prompt robustness.
    }
    \label{fig:metric_correlation}
\end{figure}

To assess whether CapTrack metrics capture distinct aspects of model behavior, we compute pairwise Spearman correlations across all sub-capabilities using the relative deviations observed across model, domain, and post-training settings. Figure~\ref{fig:metric_correlation} shows that most capability pairs exhibit weak correlations, indicating low redundancy across the evaluation suite.

Overall, 67\% of metric pairs have $|r| < 0.3$, 28\% show moderate correlation, and only 5\% are highly correlated, with a median absolute correlation of $|r|=0.21$. The few stronger correlations form interpretable clusters. For example, verbosity, formatting use, and multi-turn commitment are correlated, reflecting shared response-style shifts induced by post-training. Similarly, faithfulness and prompt robustness show related drift, suggesting that models losing calibration under domain shift may also become less stable under prompt rephrasing.

These results support the use of a capability-level evaluation suite: most metrics provide complementary information, while the observed correlations reveal meaningful relationships between specific forms of capability drift rather than broad redundancy.

\subsection{LLM-as-a-Judge Setup}
\label{app:judge}

Several CapTrack metrics rely on LLM-as-a-judge evaluation, including reasoning quality, grounding and overreach, safety compliance, behavioral preferences, and multi-turn protocol adherence. To select a judge model that balances evaluation quality and computational efficiency, we conduct a systematic comparison of candidate judges prior to running the full evaluation.

\paragraph{Judge Model Comparison.}
\begin{figure}[t]
\centering
\includegraphics[width=\textwidth]{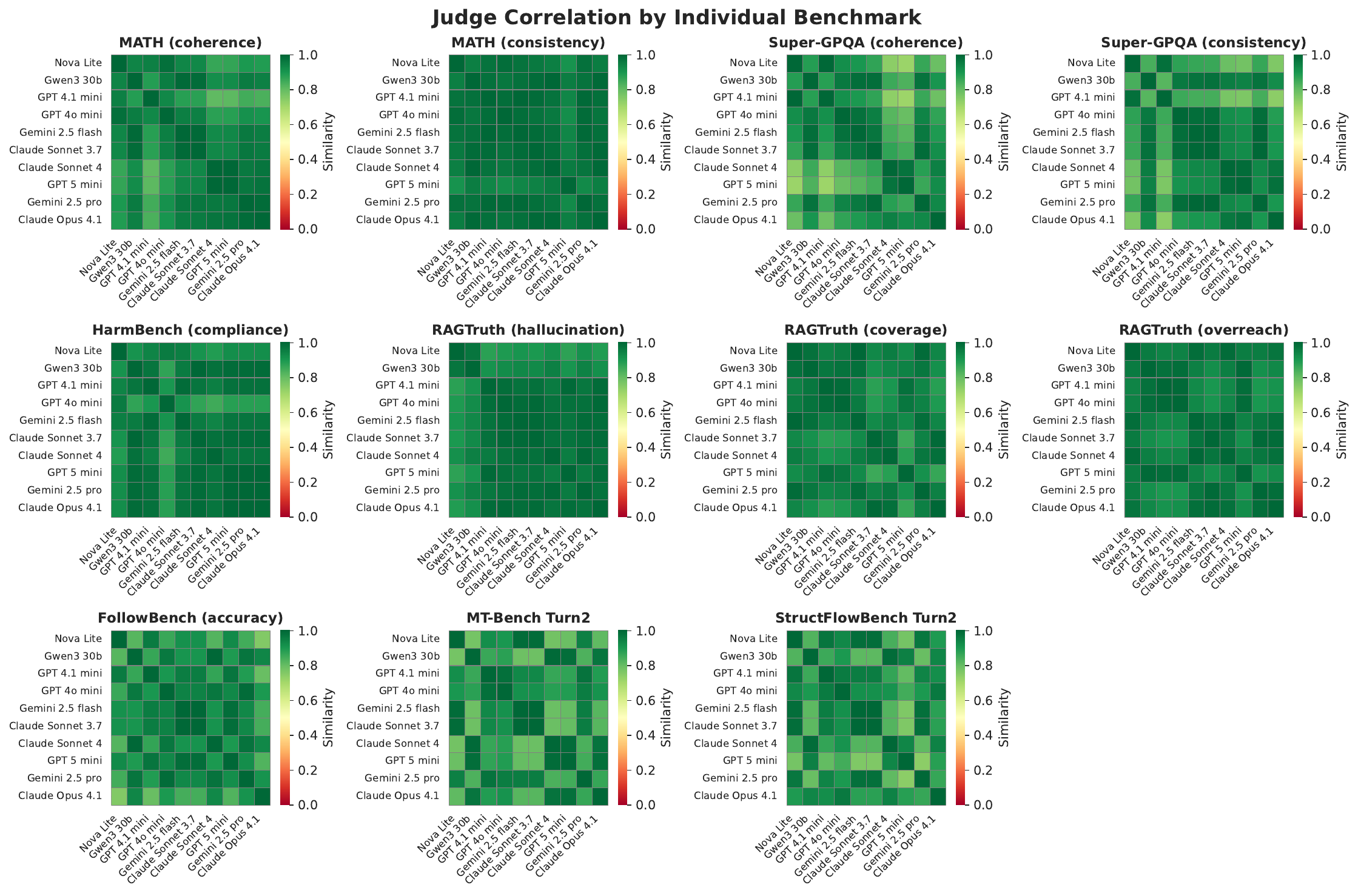}
\caption{
Pairwise correlation of LLM-as-a-judge scores across candidate judge models, computed separately for each benchmark requiring a judge. Scores are based on 100 responses per benchmark generated by the out-of-the-box \texttt{Llama-3.3-70B-Instruct} model. \texttt{GPT-4o-mini} shows consistently high correlation with strong frontier judges such as Claude Opus 4.1 and Gemini 2.5 Pro while offering substantially lower inference cost and latency, motivating its use as the default judge in CapTrack.}
\label{fig:judge-correlation}
\end{figure}

We evaluate a set of representative judge models spanning different families and sizes, including lightweight and frontier-scale models. For each benchmark requiring a judge, we sample 100 model responses generated by the out-of-the-box \texttt{Llama-3.3-70B-Instruct} model and score them using each candidate judge. We then compute pairwise correlations between the resulting judge scores, separately for each benchmark and metric.

Figure~\ref{fig:judge-correlation} reports the resulting correlation matrices across benchmarks. Strong frontier judges such as Claude Opus 4.1 and Gemini 2.5 Pro exhibit high mutual agreement, serving as a reference for evaluation quality. Importantly, \texttt{GPT-4o-mini} achieves consistently high correlations with these stronger judges across reasoning, grounding, safety, and protocol-oriented benchmarks, while offering substantially lower inference latency and cost.

\paragraph{Final Judge Choice.}
Based on the correlation analysis, we select \texttt{GPT-4o-mini} as the judge model for all LLM-as-a-judge metrics reported in this work. This choice offers a favorable trade-off between evaluation quality and computational efficiency, enabling large-scale and repeated evaluation while maintaining strong agreement with substantially more expensive judge models. To ensure consistency, we use a fixed judge configuration and prompt templates across all benchmarks, and the judge is never provided with information about the evaluated model or its post-training condition.

In addition to the quantitative agreement analysis, we perform targeted manual inspections during metric and prompt development to identify ambiguous cases, degenerate judge behavior, or systematic evaluation errors. These inspections are used exclusively to refine prompts and metric definitions and are not applied to adjust reported results post hoc.

While no automated judge can fully replace human evaluation, our analysis emphasizes relative differences between model variants evaluated under identical judging conditions. For this comparative setting, absolute judge calibration is less critical, and consistent application of the same judge suffices to support reliable conclusions.

\paragraph{Robustness of Relative Deviations.}
To assess whether our conclusions depend on the choice of judge, we evaluate the stability of relative capability deviations across a subset of representative judge models. For each judge-based metric, we recompute relative deviations using multiple judges and measure the resulting variance across judge choices.

\begin{figure}[t]
    \centering
    \includegraphics[width=0.9\linewidth]{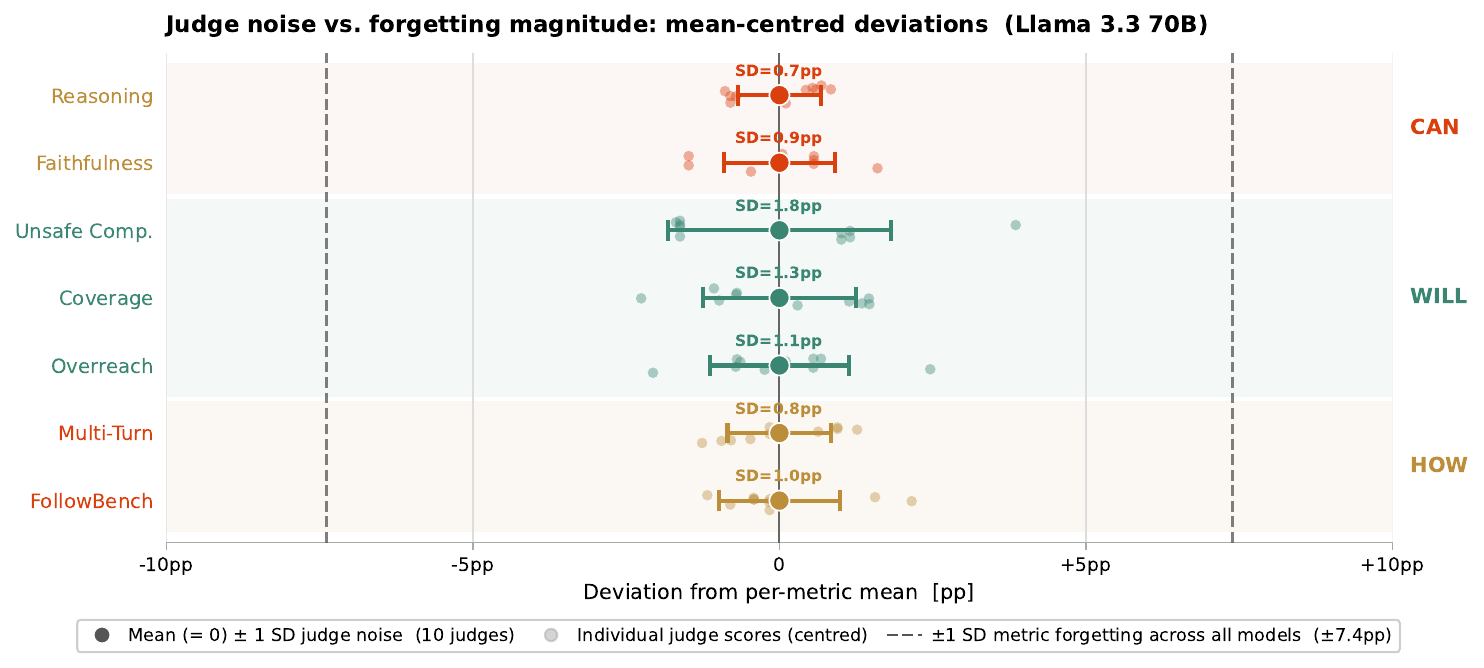}
    \caption{
    \textbf{Stability of relative deviations across judge models.}
    Relative capability deviations are recomputed using multiple LLM judges. 
    Error bars show inter-judge variability. Variance across judges is small compared to overall capability drift, indicating that CapTrack results are robust to judge selection.
    }
    \label{fig:judge-variance}
\end{figure}

We find that relative deviations are highly stable: across all evaluated settings, the mean inter-judge standard deviation is 1.1 percentage points, substantially smaller than the observed variance in capability-level drift across models and post-training methods. This indicates that while absolute scores may vary slightly across judges, the relative differences that drive our analysis remain consistent.

These results support the use of a single efficient judge model (GPT-4o-mini) for large-scale evaluation, as the conclusions drawn from CapTrack are robust to reasonable variation in judge choice.

\section{Additional Experimental Results}
\label{app:additional-results}

This appendix presents additional experimental results that complement the analyses in the main paper but could not be included due to space constraints. These results provide a more complete view of capability-level effects across post-training algorithms, domains, and model families, and are intended to support and contextualize the findings discussed in Section~\ref{sec:experiments}.

\subsection{Extended Spider Plot Results}
\label{app:spider-extended}
\begin{figure*}[t]
\centering
\includegraphics[width=0.49\textwidth]{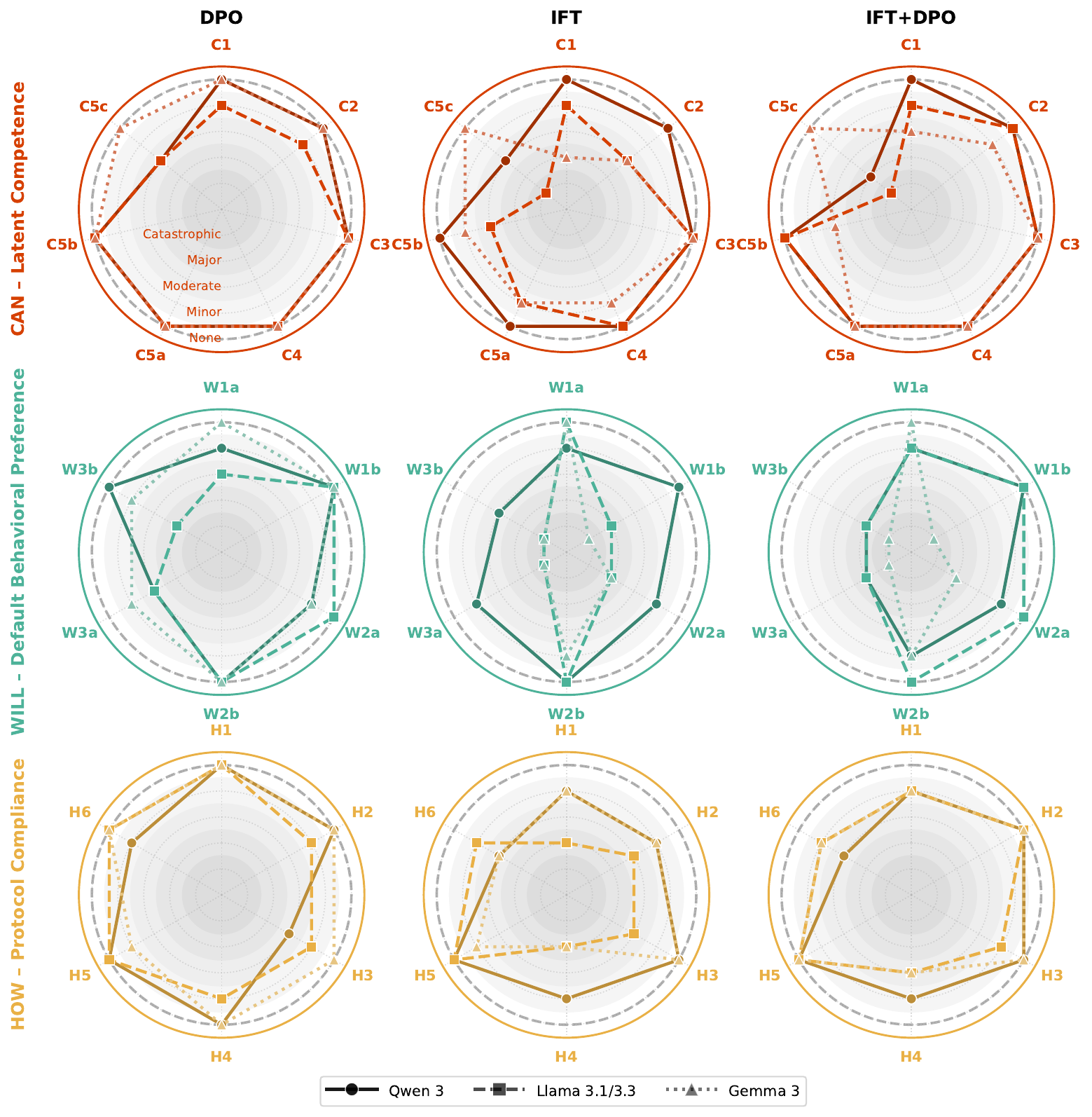}
\hfill
\includegraphics[width=0.49\textwidth]{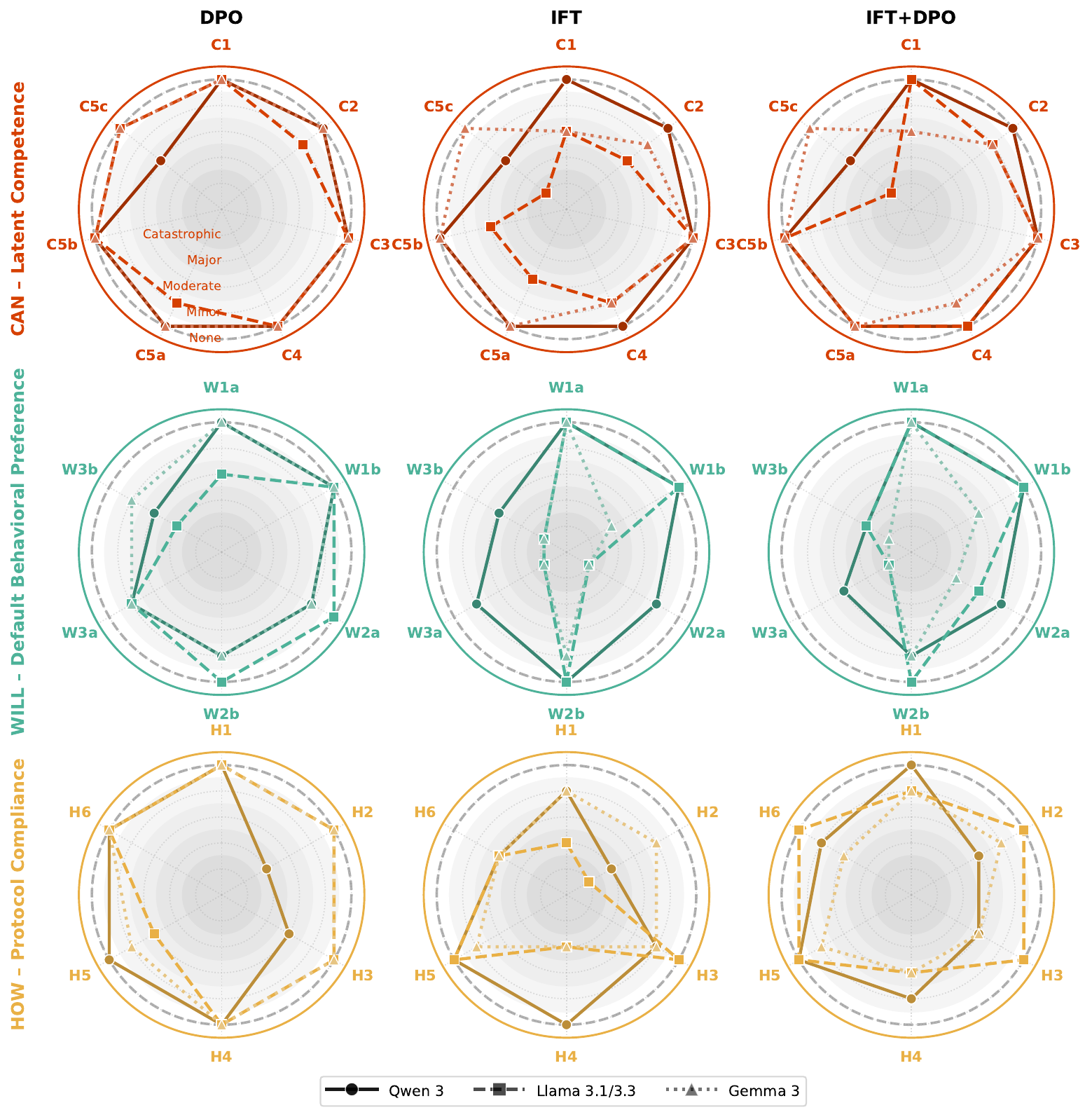}
\caption{
Extended spider plot results across model families. \textbf{Left:} Legal-domain results including DPO, IFT, and IFT+DPO. \textbf{Right:} Corresponding results for the medical domain. Each spider plot shows results averaged across model families for CAN (latent competence), WILL (default behavioral preferences), and HOW (protocol compliance), with radial distance indicating increasing forgetting. These figures complement the main paper spider plots by showing the combined post-training setting and domain-specific effects.}
\label{fig:spider-extended}
\end{figure*}

Figure~\ref{fig:spider-extended} extends the spider plot analysis from the main paper by including the missing IFT+DPO configuration as well as the corresponding results for the medical domain. As in the main paper, each spider plot summarizes capability-level changes across the CAN, WILL, and HOW categories, aggregated by model family. These additional views confirm the qualitative trends observed in the main figures and illustrate how combined post-training (IFT+DPO) and domain shifts affect different capability groups.

\subsection{Per-Model (Non-Aggregated) Results}
\label{app:per-model-results}

\begin{figure*}[t]
\centering
\includegraphics[width=\textwidth]{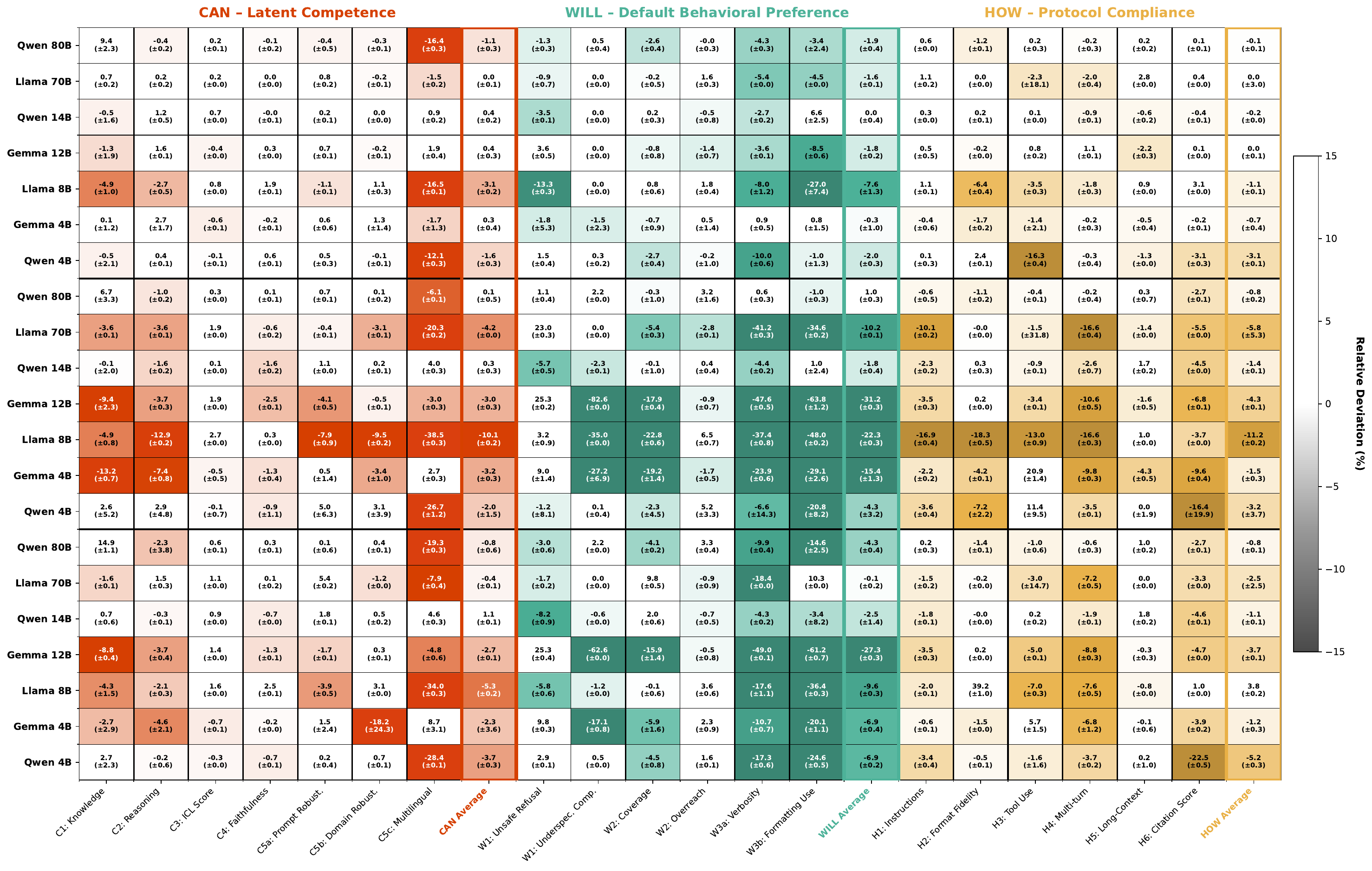}
\caption{
Per-model, non-aggregated CapTrack results for the legal domain. Each row corresponds to an individual model and post-training configuration (IFT, DPO, IFT+DPO), and each column corresponds to a CAN-, WILL-, or HOW-level metric or category average. Values denote relative deviation (\%) with respect to the corresponding base model.}
\label{fig:per-model-legal}
\end{figure*}
\begin{figure*}[t]
\centering
\includegraphics[width=\textwidth]{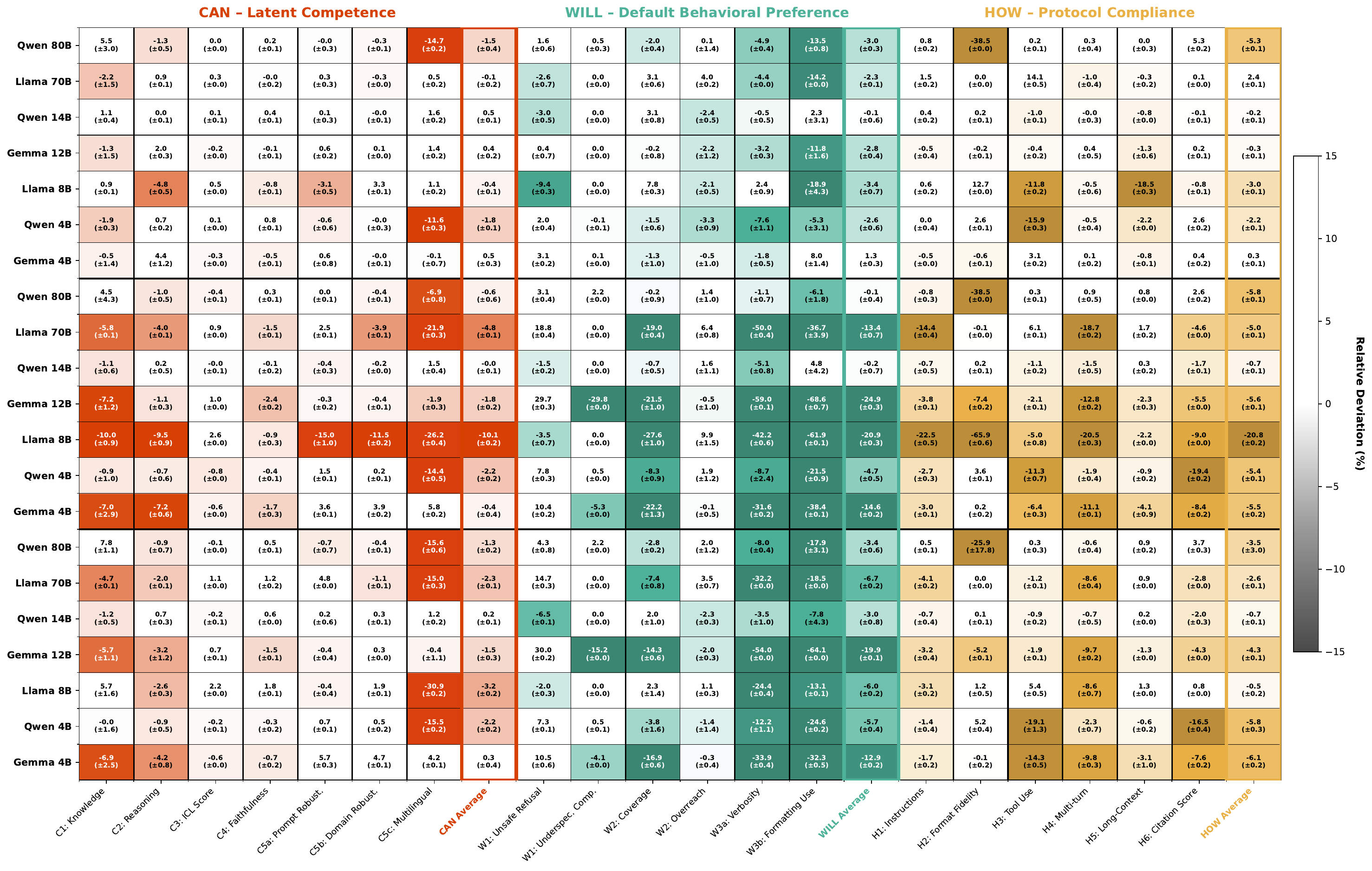}
\caption{
Per-model, non-aggregated CapTrack results for the medical domain. The layout and interpretation mirror Figure~\ref{fig:per-model-legal}, enabling direct comparison of model-specific post-training effects across domains.}
\label{fig:per-model-medical}
\end{figure*}

To complement the aggregated analyses presented in the main paper and Appendix~\ref{app:spider-extended}, we report per-model, non-aggregated results for both the legal and medical domains. These results reveal heterogeneity in post-training effects across individual models and help contextualize the variance observed within model families.

Figures~\ref{fig:per-model-legal} and~\ref{fig:per-model-medical} show detailed heatmaps of relative performance deviations for each capability dimension, broken down by model, post-training method (IFT, DPO, and IFT+DPO), and domain. Each row corresponds to a specific model and post-training configuration, while columns represent individual CAN-, WILL-, and HOW-level metrics as well as their category-level averages. Values indicate relative deviation with respect to the corresponding base model, following the conventions described in Section~\ref{sec:captrack}.

These non-aggregated views provide a more fine-grained perspective on the aggregated trends reported in the main paper, revealing how post-training effects vary across model families and parameter scales. While the overall qualitative patterns remain consistent with the family-level analysis, the per-model breakdown highlights systematic differences between architectures and sizes, particularly in how strongly individual models trade off latent competence (CAN) against behavioral (WILL) and protocol-level (HOW) properties. This additional resolution helps contextualize the aggregated results and illustrates how model-specific characteristics modulate post-training behavior.

\subsection{Forgetting by Model Size}
\begin{figure*}[t]
\centering
\includegraphics[width=\textwidth]{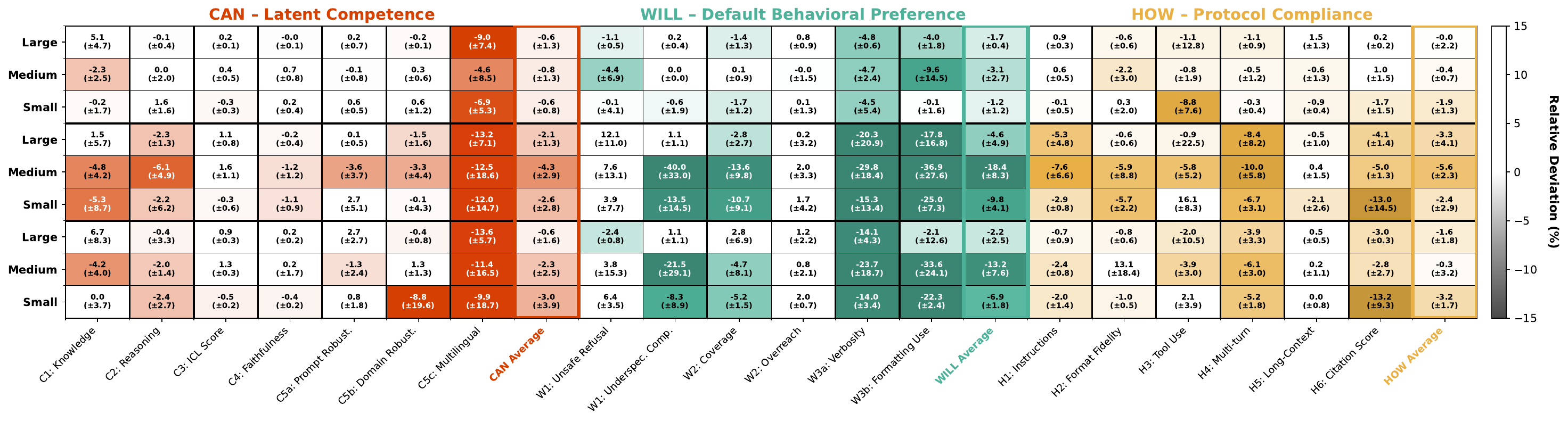}
\caption{
Capability-level forgetting aggregated by model size category (Small, Medium, Large) across model families for legal-domain post-training. Values report relative deviation (\%) with respect to the corresponding out-of-the-box model (higher is better). Results are shown for IFT, DPO, and IFT+DPO. Across most capabilities, no consistent monotonic trend with model size is observed, indicating that model scale alone does not determine forgetting behavior. Citation mechanics (H6) constitute a notable exception, where smaller models exhibit stronger degradation.}
\label{fig:forgetting-by-size}
\end{figure*}

To further disentangle the role of model scale from architectural and family-specific effects, we additionally aggregate forgetting results by model size category (Small, Medium, and Large), pooling models across families within each group. Figure~\ref{fig:forgetting-by-size} reports the resulting capability-level deviations for IFT, DPO, and IFT+DPO post-training.

Across most capabilities, we observe no consistent monotonic relationship between model size and forgetting. In particular, smaller models do not uniformly exhibit higher forgetting, nor do larger models consistently preserve capabilities better. Instead, the magnitude and direction of forgetting vary across capabilities and post-training algorithms, indicating that model size alone is not a reliable predictor of post-training stability.

Despite the absence of a consistent global trend with respect to model size, we observe that for certain capabilities, smaller models do exhibit systematically higher forgetting. The clearest instance of such a size-dependent pattern appears in citation-related behavior (H6), where smaller models show larger relative degradation. This sensitivity is plausibly attributable to the higher capacity and representational demands of citation mechanics. Outside these capability-specific effects, however, we find no monotonic relationship between model scale and forgetting, reinforcing our central conclusion that forgetting is not primarily driven by model size but rather by the interaction among post-training objectives, data, and the specific capabilities being evaluated.

\subsection{Capability-Level Deep Dives}
\label{app:captrack-deepdives}

\begin{figure*}[!t]
\centering
\includegraphics[width=\textwidth]{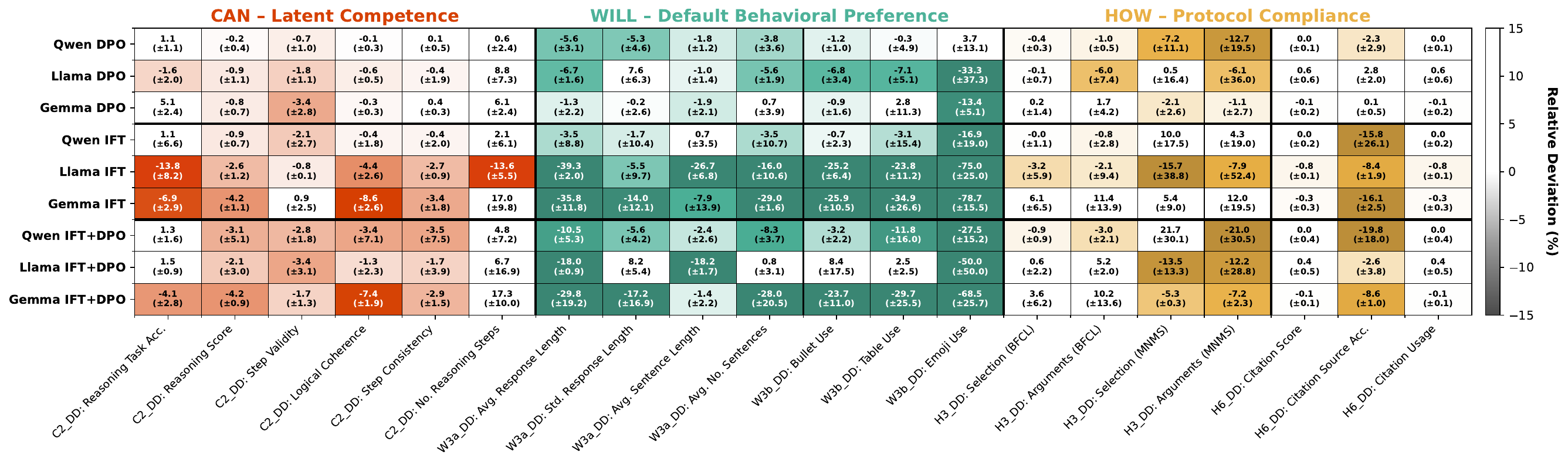}
\caption{
Capability-level deep-dive analysis using CapTrack. The figure decomposes aggregated capabilities into their constituent components: reasoning (C2\textsubscript{DD}), verbosity (W3a\textsubscript{DD}), formatting (W3b\textsubscript{DD}), tool use (H3\textsubscript{DD}), and citation mechanics (H6\textsubscript{DD}). Values report relative deviation (\%) with respect to the corresponding out-of-the-box model, averaged across benchmarks where applicable. The deep dives reveal that aggregated forgetting often arises from specific subcomponents—such as task accuracy in reasoning, variance reduction in verbosity, or source accuracy in citations—highlighting the diagnostic resolution provided by CapTrack.
}
\label{fig:captrack-deepdives}
\vspace{-10pt}
\end{figure*}

Throughout the paper, CapTrack reports aggregated capability scores that combine multiple benchmarks and metrics into stable, interpretable summaries. For example, reasoning performance (C2) aggregates task accuracy and a composite reasoning score (capturing step validity, logical coherence, and step consistency) across MATH and SuperGPQA. This aggregation is essential for drawing robust high-level conclusions about forgetting patterns across models, domains, and post-training algorithms.

At the same time, CapTrack retains the full low-level structure of its evaluation suite, enabling targeted drill-down analyses when a more fine-grained diagnosis is required. Such deep dives allow us to decompose aggregated capability shifts into their constituent components and to identify which specific aspects of a capability are most affected by post-training.

Figure~\ref{fig:captrack-deepdives} presents several illustrative deep dives. For reasoning (C2\textsubscript{DD}), we report individual components averaged across benchmarks, including task accuracy, reasoning score, step validity, logical coherence, step consistency, and the number of reasoning steps. For default behavioral preferences related to verbosity (W3a\textsubscript{DD}) and formatting (W3b\textsubscript{DD}), we decompose the aggregated scores into response length statistics (mean and standard deviation, sentence length, number of sentences) and formatting usage rates (bullets, tables, emojis). For tool use (H3\textsubscript{DD}), we report selection and argument accuracy separately for BFCL and the more challenging multi-turn MNMS benchmark. Finally, for citation mechanics (H6\textsubscript{DD}), we distinguish between citation usage, format accuracy, and source accuracy on HotpotQA.

These deep dives reveal several consistent patterns. For reasoning, forgetting is primarily driven by decreases in task accuracy, while degradation in reasoning quality is most pronounced in logical coherence and step consistency. In addition, for Llama models, IFT leads to a reduction in the number of explicit reasoning steps, suggesting a shift toward shorter or more implicit reasoning traces. For verbosity, reductions in average response length can be attributed jointly to shorter sentences and fewer sentences per response; moreover, IFT consistently reduces the variance of response length, leading to more uniform outputs. For formatting, usage rates decline across Llama and Gemma models after IFT, with emoji usage showing the most consistent and pronounced drop across families and post-training configurations. In tool use, forgetting is concentrated on the harder multi-turn MNMS benchmark, whereas single-turn BFCL performance remains comparatively stable. For citations, degradation is largely confined to source accuracy, while citation usage and format correctness are relatively unaffected.

Taken together, these analyses highlight the diagnostic value of CapTrack beyond aggregated scores. While high-level metrics are sufficient to characterize overall forgetting trends, the ability to drill down into individual components reveals which mechanisms drive observed capability shifts and where mitigation efforts may need to be targeted. This flexibility underscores CapTrack’s usefulness as a comprehensive evaluation suite for analyzing the multifaceted and heterogeneous nature of post-training effects.

\subsection{Additional Post-Training Algorithm: GRPO}
\label{app:grpo}

To extend our analysis beyond supervised fine-tuning (IFT) and preference optimization (DPO), we include preliminary experiments with Group Relative Policy Optimization (GRPO) \citep{shao2024deepseekmathpushinglimitsmathematical}, a reinforcement learning–based post-training method that optimizes model behavior with respect to a learned reward signal.

\paragraph{Setup.}
We evaluate GRPO on the legal domain using Qwen~3~14B, Llama~3.1~8B, and Gemma~3~12B. Models are trained for a single epoch on $\sim$41k legal samples with a domain-specific reward model. Training configurations follow the same general setup as IFT and DPO (Appendix~A.3).

\begin{figure}[!t]
    \centering
    \includegraphics[width=0.7\linewidth]{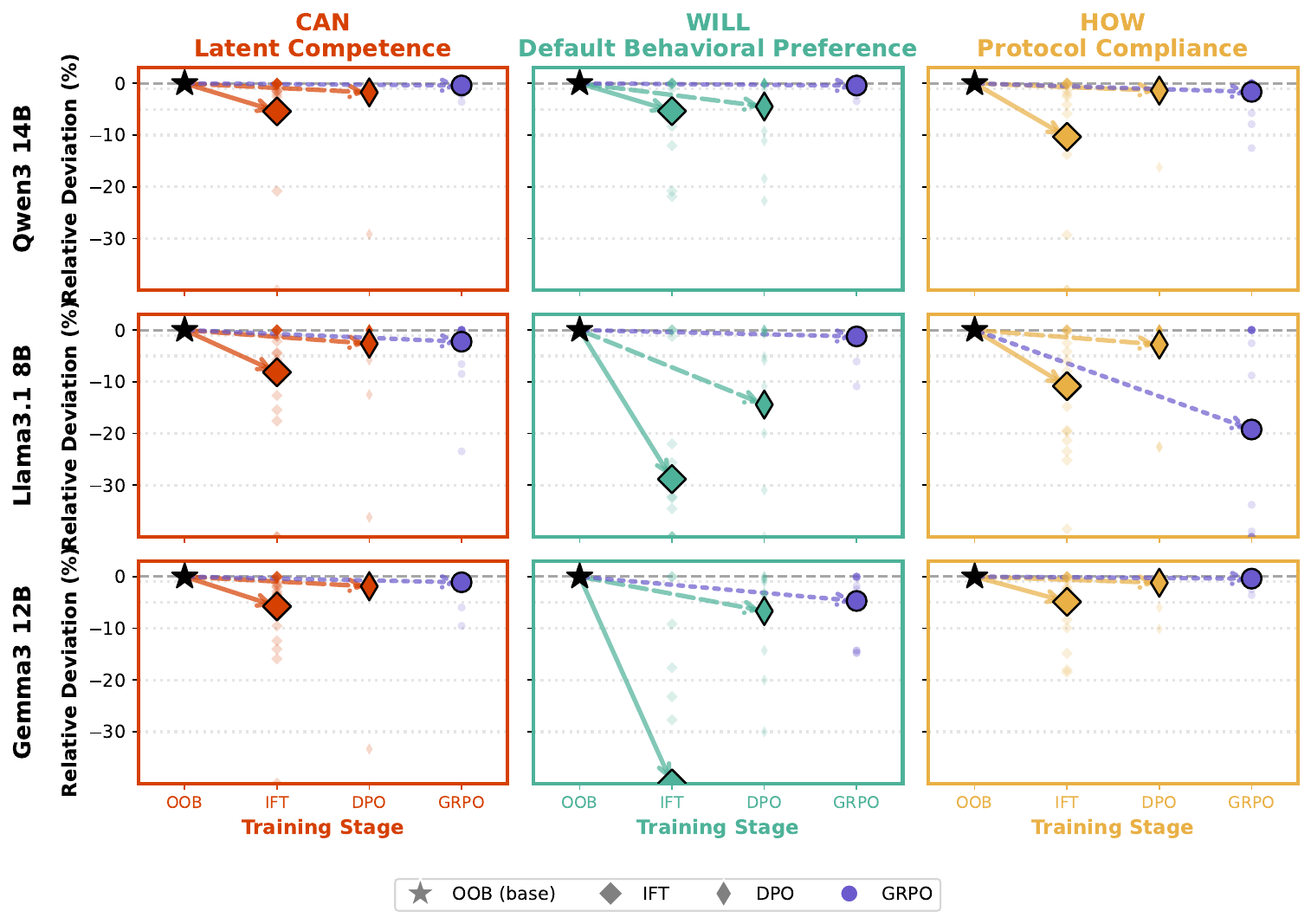}
    \caption{
    \textbf{Capability-level forgetting under GRPO on the legal domain.}
    Relative deviation from the OOB model is shown for CAN (latent competence), WILL (behavioral preferences), and HOW (protocol compliance), aggregated across benchmarks. 
    GRPO exhibits moderate overall forgetting comparable to DPO, with minor degradation in CAN and WILL, and more pronounced drops in HOW, particularly for instruction following and format fidelity.
    }
    \label{fig:grpo_forgetting}
\end{figure}

\paragraph{Results.}
Figure~\ref{fig:grpo_forgetting} shows capability-level forgetting under GRPO, aggregated across benchmarks and reported as relative deviation from the OOB model. Overall, GRPO exhibits moderate forgetting, with average drift comparable to DPO (e.g., $\sim$-4\% across capability groups).

At the capability level, GRPO largely preserves latent competence (CAN) and behavioral preferences (WILL), showing only minor degradation. In contrast, for the Llama model, protocol-level capabilities (HOW) exhibit more pronounced declines, particularly in instruction-following and output-format fidelity.

\paragraph{Discussion.}
These results indicate that GRPO behaves similarly to DPO in terms of overall stability, supporting the view that optimization methods with implicit or explicit regularization toward a reference policy induce less capability drift than unconstrained supervised fine-tuning. However, the observed degradation in protocol-level capabilities suggests that reward design and optimization dynamics can selectively impact execution reliability.

Overall, GRPO reinforces the broader conclusion that post-training-induced forgetting is algorithm-dependent and manifests heterogeneously across capability groups.

\subsection{Interpreting Behavioral Drift vs.\ Capability Loss}
\label{app:interpreting_drift}

CapTrack is designed to surface \emph{relative capability and behavioral drift} induced by post-training, rather than to impose a normative notion of model quality. In particular, changes observed in the \textsc{WILL} capability group reflect shifts in a model’s \emph{default response behavior}, which may be beneficial, detrimental, or ambiguous depending on the deployment context. We therefore emphasize that not all measured drift should be interpreted as inherently undesirable.

Several behavioral changes observed in our experiments illustrate this distinction. For example, reductions in verbosity and formatting diversity following instruction fine-tuning can be interpreted as a loss of expressiveness or user engagement in open-ended settings, but may be advantageous in domains that favor concise, standardized responses. Similarly, increased refusal rates under underspecified prompts may reduce usability in exploratory interactions, while simultaneously reflecting more conservative or safety-oriented default policies.

Importantly, CapTrack distinguishes such behavioral shifts from losses in latent competence. In many cases, models exhibiting strong drift in \textsc{WILL} retain stable performance under \textsc{CAN} when prompted appropriately, indicating that underlying capabilities remain intact and that changes are primarily driven by altered response preferences rather than forgetting of skills or knowledge.

By reporting capability-level deviations relative to an out-of-the-box baseline, CapTrack makes these trade-offs explicit without prescribing how they should be resolved. We view this diagnostic perspective as a strength: rather than collapsing diverse post-training effects into a single performance score, CapTrack enables practitioners to assess which dimensions of behavior change, and to judge their desirability in light of application-specific requirements.

Overall, CapTrack should be understood as a tool for exposing and comparing post-training-induced trade-offs across competence, behavior, and execution. Determining whether a given form of drift constitutes harmful forgetting, acceptable alignment, or intentional specialization ultimately requires domain knowledge and deployment-specific evaluation, which are beyond the scope of a general-purpose benchmark.

\subsection{In-Scope Domain Performance}
\label{app:in-scope}

In addition to the capability-oriented evaluations reported throughout the paper, we assess model behavior on in-scope benchmarks drawn from the target post-training domains. These evaluations are intended as sanity checks to verify that post-training yields meaningful domain adaptation under realistic constraints, rather than as comprehensive measures of downstream performance. In particular, they are not designed to assess generalization or robustness, nor do they serve as the primary basis for our forgetting analysis.

\paragraph{Legal Domain.}
For the legal domain, we evaluate models on the Stanford \emph{LegalBench} benchmark \citep{guha2023legalbenchcollaborativelybuiltbenchmark}, which comprises a heterogeneous collection of legal reasoning and classification tasks spanning statutory interpretation, contract analysis, and case-law understanding. Performance is measured using the benchmark’s task-specific accuracy metrics and reported as the average performance across all LegalBench tasks, relative to the corresponding out-of-the-box (OOB) model.

\paragraph{Medical Domain.}
For the medical domain, we consider three complementary evaluation settings. First, we evaluate models on the \emph{Medical Meadow Health Advice} test set \citep{yu-etal-2019-detecting}, which probes a model’s ability to appropriately calibrate medical guidance by classifying responses as providing strong advice, weak advice, or no advice.

Second, we evaluate models on the test set of the \emph{PubMed Causal} benchmark, which assesses whether a model can correctly identify the nature of relationships described in biomedical text. Given a short passage, the model is tasked with classifying the relationship as directly correlative, conditionally causative, causative, or indicating no relationship. Performance is measured using classification accuracy and reported relative to the corresponding OOB model.

Third, we assess preference-level improvements using the \emph{AskDocs} test split \citep{li2025aligningllmsaskgood}. For each prompt, we generate responses from both the post-trained model and its OOB counterpart and prompt an LLM judge (\texttt{gpt-4o-mini}) to perform a blind pairwise comparison. Results are reported as win rates (\%) of the post-trained model against the OOB baseline. As the preference signal for medical advice quality is explicitly encoded in the DPO training mixture, improvements on AskDocs are primarily expected for post-training configurations involving DPO; consequently, IFT-only post-training does not consistently improve performance on this benchmark.

\begin{table*}[t]
\centering
\small
\resizebox{\textwidth}{!}{%
\begin{tabular}{l l l c c c c c c c}
\toprule
\textbf{Domain} & \textbf{Benchmark} & \textbf{Alg.} &
\textbf{Qwen-80B} & \textbf{Qwen-14B} & \textbf{Qwen-4B} &
\textbf{LLaMA-70B} & \textbf{LLaMA-8B} &
\textbf{Gemma-12B} & \textbf{Gemma-4B} \\
\midrule

\multirow{3}{*}{Legal}
& \multirow{3}{*}{LegalBench}
& DPO     & $+0.09$ & $-0.03$ & $+5.80$ & $+4.93$ & $+15.35$ & $+0.64$ & $+3.16$ \\
&  & IFT     & $-0.69$ & $+1.23$ & $+3.21$ & $+0.00$ & $+9.55$  & $+1.51$ & $+1.18$ \\
&  & IFT+DPO & $-0.79$ & $+1.44$ & $+7.50$ & $+3.98$ & $+18.94$ & $+1.32$ & $+1.18$ \\
\midrule

\multirow{9}{*}{Medical}
& \multirow{3}{*}{HealthAdvice}
& DPO     & $+5.20$ & $+3.11$ & $+2.23$ & $+11.94$ & $+6.84$ & $+0.73$ & $+1.96$ \\
&  & IFT     & $+7.54$ & $+1.56$ & $-10.63$ & $+140.26$ & $-33.80$ & $+3.97$ & $+3.55$ \\
&  & IFT+DPO & $+9.24$ & $+6.05$ & $-7.60$  & $+139.13$ & $+35.26$ & $+3.59$ & $+4.07$ \\
\cmidrule{2-10}
& \multirow{3}{*}{PubMed Causal}
& DPO     & $+3.97$ & $-17.06$ & $+0.00$ & $-44.71$ & $+14.29$ & $+60.56$ & $+175.00$ \\
&  & IFT     & $+39.98$ & $+14.64$ & $+25.00$ & $+34.20$ & $+17.48$ & $+63.56$ & $+349.80$ \\
&  & IFT+DPO & $+51.98$ & $+2.48$  & $+25.00$ & $+26.31$ & $-49.21$ & $+63.56$ & $+341.53$ \\
\cmidrule{2-10}
& \multirow{3}{*}{AskDocs}
& DPO     & 68.8 & 85.9 & 79.2 & 81.1 & 88.5 & 78.2 & 78.2 \\
&  & IFT     & 42.2 & 30.4 & 35.1 & 36.7 & 8.9  & 13.7 & 13.4 \\
&  & IFT+DPO & 81.6 & 86.2 & 85.6 & 84.5 & 86.5 & 71.7 & 59.2 \\
\bottomrule
\end{tabular}
}
\caption{
In-scope domain performance for legal and medical benchmarks. For LegalBench, HealthAdvice, and PubMed Causal, values report relative deviation (\%) with respect to the corresponding out-of-the-box (OOB) model (higher is better). For AskDocs, values are win rates (\%) from a blind pairwise LLM-judge comparison of the post-trained model against its OOB counterpart.}
\label{tab:in-scope-results}
\end{table*}

\paragraph{In-Scope Performance Overview.}
Across domains, models, and post-training configurations, the in-scope evaluations show that post-training generally leads to improved or competitive performance on domain-relevant benchmarks, indicating non-degenerate and effective domain adaptation. At the same time, we observe heterogeneity in the magnitude of gains, with some large or already strong OOB models exhibiting small improvements or near-neutral changes on individual public benchmarks. Such variability is expected in realistic post-training settings, where publicly available benchmarks may only partially reflect the objectives and distributions targeted during adaptation.

Importantly, our analysis of forgetting does not rely on uniform or maximal in-domain performance gains. Rather, it focuses on how different post-training procedures reshape model capabilities relative to their OOB counterparts under comparable optimization effort. The presence of capability-level forgetting alongside largely successful domain adaptation underscores that forgetting is not merely a symptom of failed post-training, but a systematic and consequential byproduct of effective post-training itself.

\section{Extended Data-Centric Mitigation Analyses}
\label{app:data-ablations}

This appendix presents extended results on the data mitigation experiments presented in Section ~\ref{sec:data_ablation}. 

\subsection{Data Source Ablation}
\label{app:data-source-ablation}

In this ablation study, we analyze the impact of the post-training data source on capability-level forgetting by comparing models trained on a general-purpose instruction mixture (Tulu~3) with models trained on a domain-specific legal mixture. Across all settings, the post-training algorithm and overall training budget are held fixed. This experiment is designed to isolate the contribution of the data source to forgetting behavior, complementing the aggregated analysis presented in the main paper.

\begin{figure*}[t]
\centering
\includegraphics[width=\textwidth]{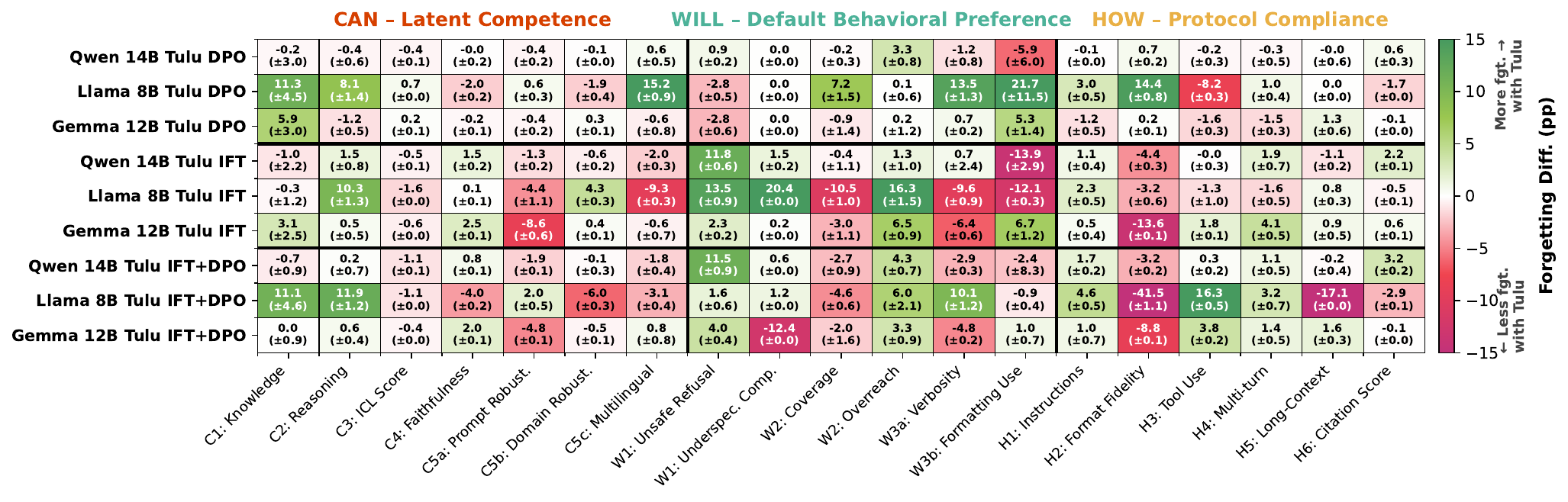}
\caption{
Data source ablation comparing post-training on a general-purpose instruction mixture (Tulu) versus a legal-domain mixture. Values show the difference in forgetting (percentage points) between Tulu-based and legal-domain post-training across capability dimensions, models, and post-training algorithms. Positive values (green) indicate less forgetting when trained on Tulu, while negative values (red) indicate less forgetting when trained on legal-domain data.}
\label{fig:tulu-vs-legal}
\end{figure*}

We report the full, non-aggregated results across model families, post-training methods (IFT, DPO, and IFT+DPO), and capability dimensions. Figure~\ref{fig:tulu-vs-legal} shows the difference in forgetting between Tulu-based and legal-domain post-training for each capability. Positive values indicate less forgetting when training on Tulu, while negative values indicate less forgetting when training on legal-domain data.

Consistent with the main-paper analysis, we find no single data source that uniformly dominates across capabilities, and, importantly, no evidence that forgetting is inherently driven by domain-specific post-training. While the aggregated results highlight systematic differences between general-purpose and domain-specific data mixtures, the per-model breakdown presented here further reveals substantial heterogeneity across models, post-training algorithms, and capability dimensions. For some model–capability combinations, post-training on Tulu reduces forgetting relative to legal-domain data, whereas for others it increases forgetting. These variations correlate with model size, architecture, and training objective, indicating that forgetting arises from interactions between the post-training procedure and the underlying model rather than from domain specificity per se.

Taken together, these results reinforce the conclusion that forgetting is a general property of post-training, rather than a consequence of adapting models to specialized domains. While data source choice remains a first-order design decision, its effects are nuanced and do not yield a universally optimal configuration. General-purpose instruction data can mitigate forgetting for certain capabilities and models, but may exacerbate it for others, underscoring the need for capability-aware post-training strategies rather than treating domain-specific data as the primary cause of forgetting.

\subsection{Data Mixture Ablation (Replay vs.\ No Replay)}
\label{app:data-mixture-ablation}

In this ablation study, we analyze the effect of removing general replay data from the legal post-training mixtures. Specifically, we compare post-training configurations that include both domain-specific and general (replay) data (\textbf{RP}) against configurations that use only domain-specific data without replay (\textbf{NRP}). As in Appendix~\ref{app:data-source-ablation}, the goal is to isolate the contribution of the data mixture composition while holding the post-training algorithm and overall training budget fixed.

We report differences in forgetting between RP and NRP configurations across capability dimensions, models, and post-training algorithms. Positive values indicate increased forgetting when replay data is included, while negative values indicate reduced forgetting.
\begin{figure*}[t]
\centering
\includegraphics[width=\textwidth]{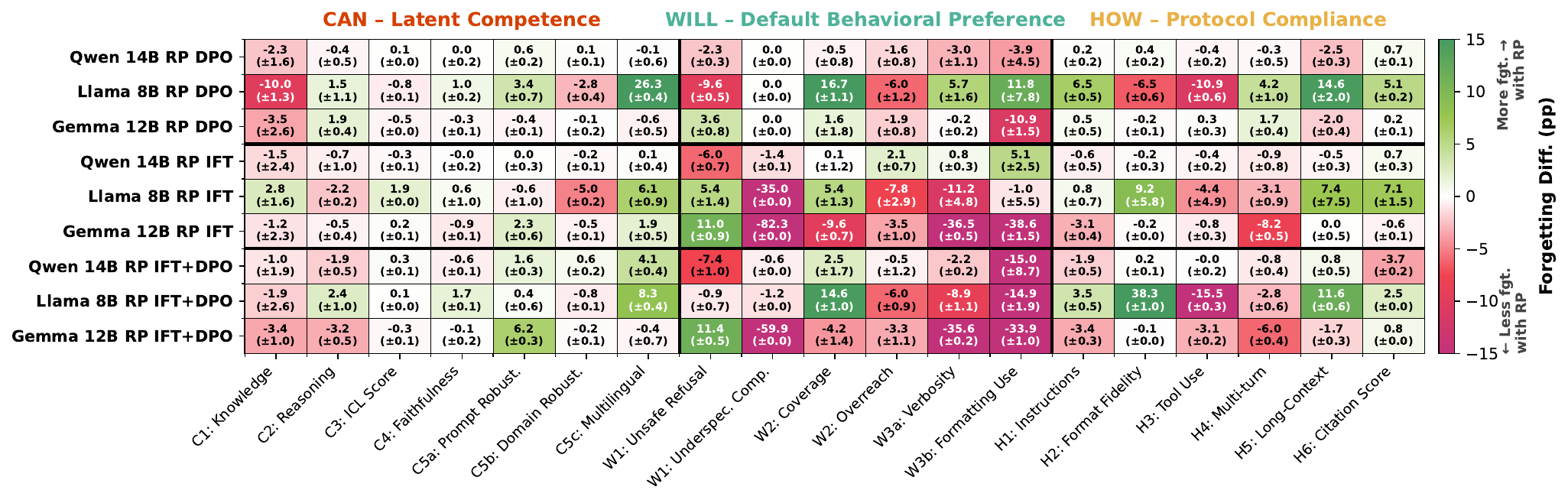}
\caption{
Non-aggregated replay ablation results across models and post-training configurations. Values show forgetting differences (percentage points) between replay (RP) and no-replay (NRP) mixtures for each capability. Positive values (red) indicate increased forgetting with replay, while negative values (green) indicate reduced forgetting with replay.}
\label{fig:rp-vs-nrp-heatmap}
\end{figure*}
\begin{wrapfigure}{r}{0.5\textwidth}
\vspace{-1.2em}
\centering
\includegraphics[width=0.48\textwidth]{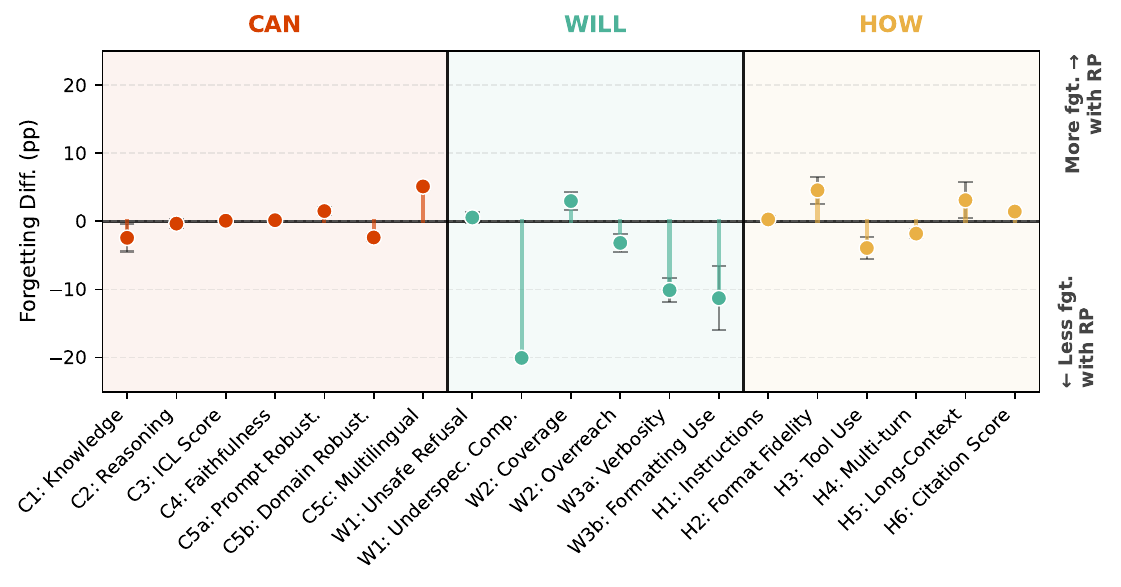}
\caption{Forgetting differences between replay (RP) and no-replay (NRP) data mixtures, averaged across models and post-training methods. Positive values indicate more forgetting with replay; negative values indicate less forgetting with replay.}
\label{fig:rp-vs-nrp-lollipop}
\vspace{-0.8em}
\end{wrapfigure}
Overall, the impact of replay data on forgetting is mixed and varies across models, capabilities, and post-training algorithms. While replay reduces forgetting for certain capabilities and model configurations, it increases forgetting for others, particularly in behavioral (WILL) and protocol-level (HOW) dimensions. These results mirror the findings of the data source ablation in Appendix~\ref{app:data-source-ablation} and suggest that incorporating general replay data does not uniformly mitigate forgetting across capabilities. Instead, its effect depends on the interaction between model characteristics, training objective, and capability type, underscoring the importance of carefully tailoring data mixtures for post-training.

\section{Extended Architectural Mitigation Analyses}
\label{app:mitigation-merging}

This appendix presents additional results on architectural mitigation via model merging. All experiments in this section are conducted on \texttt{Llama-3.1-8B-Instruct} under legal-domain post-training. As in the main paper (see Section~\ref{sec:regularization}), we quantify (i) \emph{stability} as capability retention relative to the out-of-the-box (OOB) model (higher is better), and (ii) \emph{plasticity} as in-domain legal performance relative to the corresponding unregularized post-trained baseline.

\subsection{Different Merging Approaches}
\label{app:merging-methods}

\begin{figure*}[t]
\centering
\includegraphics[width=0.8\textwidth]{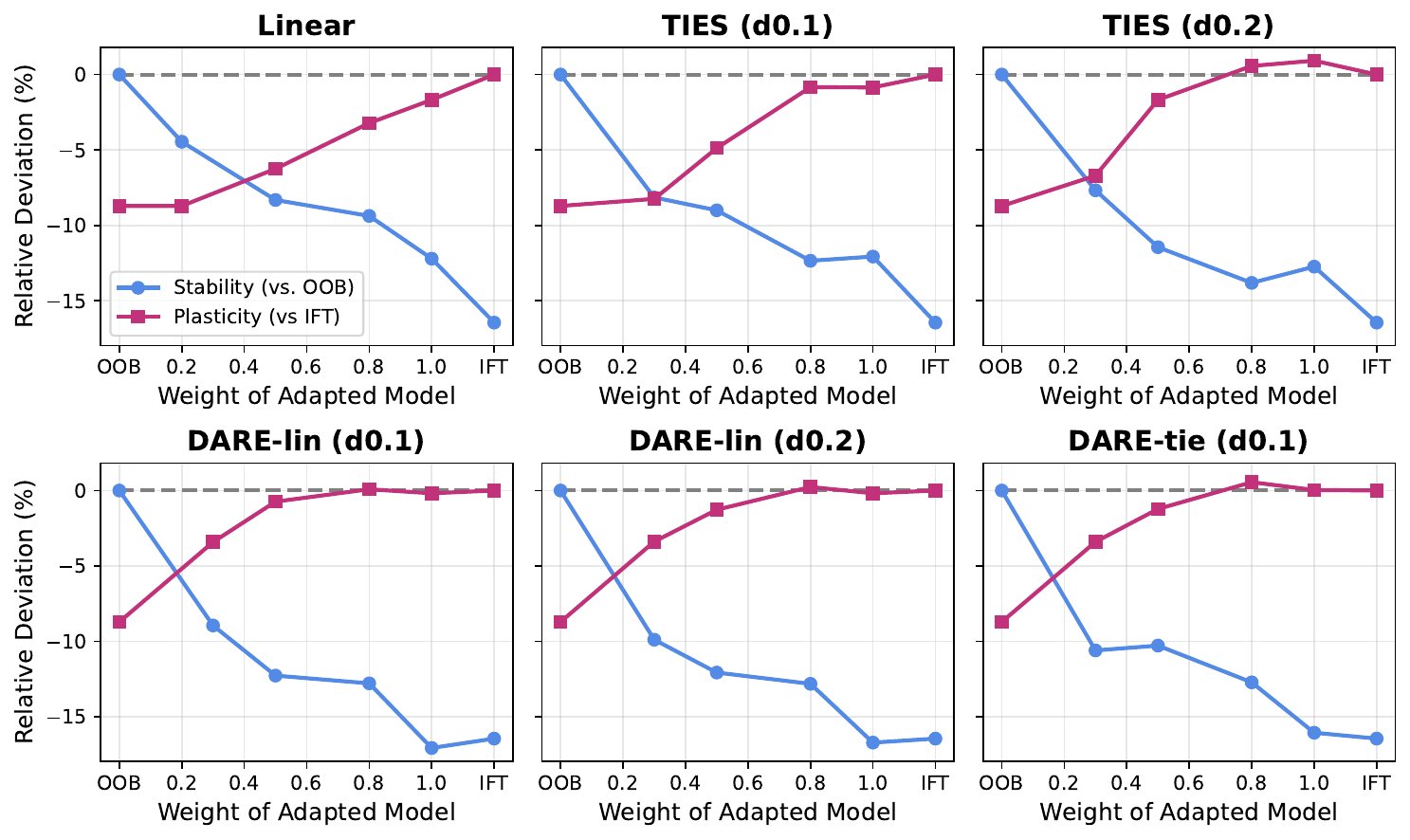}
\caption{
Stability--plasticity trade-offs for different merging operators on \texttt{Llama-3.1-8B-Instruct} with legal IFT. We compare linear interpolation, TIES (density 0.1/0.2), DARE-linear (density 0.1/0.2), and DARE-TIES (density 0.1). The x-axis denotes the weight assigned to the adapted model. Stability is measured relative to the OOB model, while plasticity is measured relative to the unregularized IFT model. Across methods, stronger regularization improves stability but reduces plasticity.}
\label{fig:merging-methods}
\end{figure*}

We compare several commonly used merging approaches to test whether specific merging operators can mitigate forgetting more effectively than the TIES configuration used in the main paper. Concretely, we evaluate: (i) linear interpolation \citep{wortsman2022modelsoupsaveragingweights}, (ii) TIES \citep{yadav2023tiesmergingresolvinginterferencemerging} with densities 0.1 and 0.2, (iii) DARE-linear \citep{yu2024languagemodelssupermario} with densities 0.1 and 0.2, and (iv) DARE-TIES \citep{yu2024languagemodelssupermario} with density 0.1. For each method, we interpolate between the OOB model and the instruction fine-tuned (IFT) model using different merge weights, and compute stability and plasticity for each merged checkpoint.

Figure~\ref{fig:merging-methods} reports stability-plasticity curves across merging methods. Across all methods, we observe a consistent stability-plasticity trade-off: increasing reliance on the adapted model improves in-domain performance but increases capability drift, while stronger regularization toward the OOB model preserves stability at the expense of plasticity. Differences between merging operators primarily affect the slope and curvature of this trade-off, but none of the tested methods eliminates it. These results underline that general-purpose merging operators are unlikely to provide a ``free lunch'' solution for mitigating capability-level forgetting without limiting learning, and that more selective, capability-aware interventions may be required.

\subsection{Merging for DPO and IFT+DPO}
\label{app:merging-dpo-iftdpo}

\begin{figure*}[t]
\centering
\includegraphics[width=\textwidth]{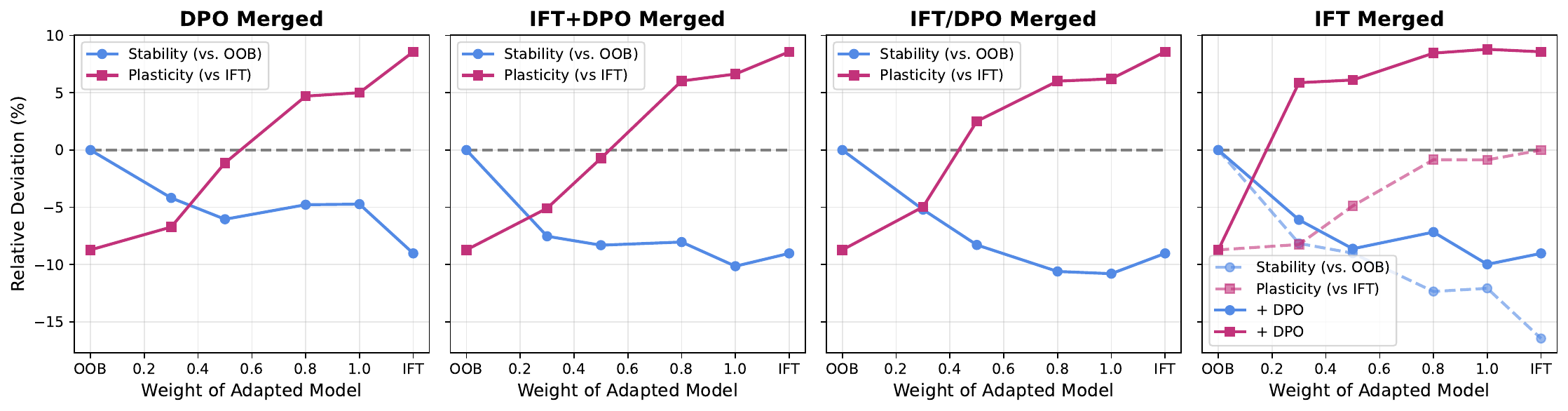}
\caption{
Stability--plasticity trade-offs for merging under DPO and sequential IFT+DPO post-training on \texttt{Llama-3.1-8B-Instruct} (legal domain). We compare: merging DPO with OOB (\emph{DPO merged}); merging the final sequential checkpoint with OOB (\emph{IFT+DPO merged}); jointly merging OOB, IFT, and DPO (\emph{IFT/DPO merged}); and applying DPO on top of an IFT-merged model (\emph{IFT merged + DPO}). The x-axis denotes the weight assigned to the adapted model. Stability is measured relative to OOB; plasticity is measured relative to the unregularized IFT model. Across all variants, stronger regularization improves stability but reduces plasticity; applying DPO after merging can partially recover plasticity but does not remove the trade-off.}

\label{fig:merging-dpo-iftdpo}
\end{figure*}

To complement the IFT-focused merging results, we additionally study model merging under (i) DPO-only post-training and (ii) sequential IFT+DPO post-training. For DPO, we merge the DPO-adapted model with the corresponding OOB model, as in the IFT setting. For sequential IFT+DPO, merging can be applied at different stages, yielding multiple variants:
(i) \emph{IFT+DPO merged}, where the final IFT+DPO checkpoint is merged with the OOB model;
(ii) \emph{IFT/DPO merged}, where we merge three checkpoints jointly (OOB, IFT, and DPO);
and (iii) \emph{IFT merged + DPO}, where we first merge the IFT model with OOB and then apply DPO on top of the merged model.

As in the main paper, we report stability relative to the OOB model. For plasticity, we report in-domain performance relative to the unregularized IFT baseline (including in the IFT+DPO variants), to maintain comparability with the IFT-centered notion of adaptation.

Figure~\ref{fig:merging-dpo-iftdpo} summarizes the resulting trade-offs. Merging for DPO exhibits a pattern similar to IFT: increasing the weight on the adapted model improves plasticity but reduces stability. For sequential IFT+DPO, all three variants again follow the same qualitative trade-off, indicating that merging does not fundamentally change the stability-plasticity tension. However, \emph{IFT merged + DPO} can partially recover plasticity while retaining some of the stability gains from merging, suggesting that preference optimization on top of a merged IFT model can mitigate forgetting to some extent. Nonetheless, the overall results show that this is not a complete solution: improvements remain constrained by the same global trade-off, and no configuration selectively protects the most vulnerable capabilities without reducing adaptation.

\section{Extended Regularization-Based Mitigation Analyses}
\label{app:lora}

This appendix presents additional results for regularization-based mitigation via Low-Rank Adaptation (LoRA) \citep{hu2021loralowrankadaptationlarge}. While the main paper analyzes the effect of varying the LoRA rank, we here extend this analysis by studying the impact of the learning rate, providing further insight into how optimization strength influences the stability-plasticity trade-off.

\subsection{Learning Rate Sensitivity}
\label{app:lora-lr}

\begin{figure*}[t]
\centering
\includegraphics[width=\textwidth]{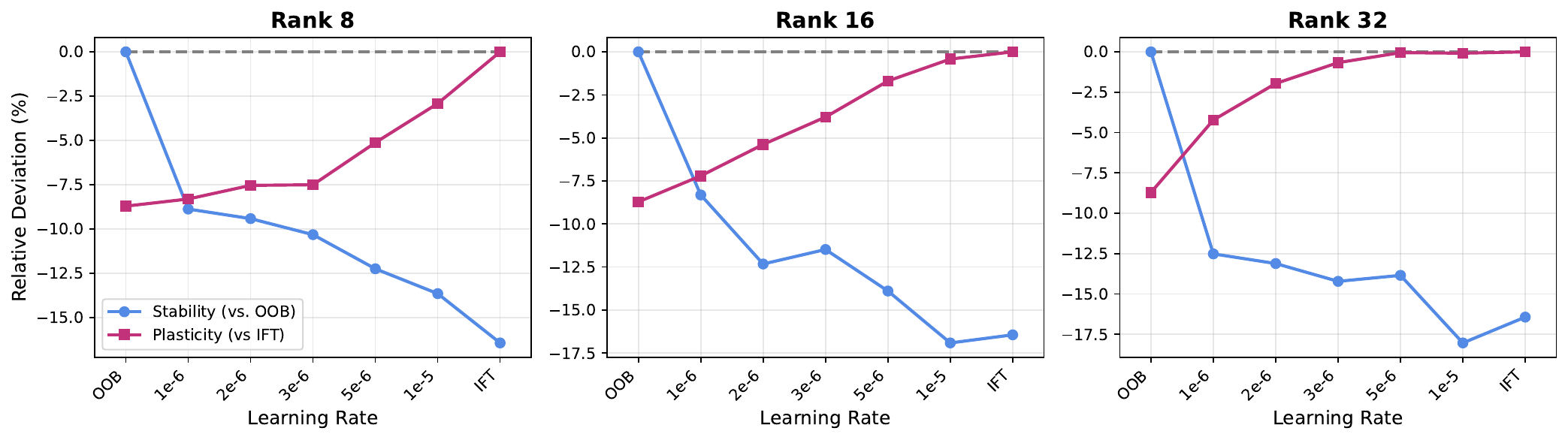}
\caption{
Stability and plasticity as a function of the learning rate for LoRA-based post-training on \texttt{Llama-3.1-8B-Instruct} (legal domain), shown for LoRA ranks 8, 16, and 32. Stability is measured as relative deviation with respect to the out-of-the-box (OOB) model, while plasticity is measured relative to the unregularized IFT baseline. Across all ranks, increasing the learning rate consistently improves plasticity while reducing stability, yielding a trade-off analogous to that observed when increasing the LoRA rank.}
\label{fig:lora-lr}
\end{figure*}

We evaluate LoRA-based post-training across a range of learning rates from $10^{-6}$ to $10^{-5}$, while keeping all other training settings fixed. Experiments are conducted on \texttt{Llama-3.1-8B-Instruct} with legal-domain post-training, and we report stability relative to the out-of-the-box (OOB) model and plasticity relative to the unregularized IFT baseline, consistent with the definitions used throughout the paper.

Figure~\ref{fig:lora-lr} shows stability and plasticity as a function of the learning rate for three LoRA ranks (8, 16, and 32). Across all ranks, increasing the learning rate consistently improves plasticity while reducing stability, yielding a trade-off that closely mirrors the behavior observed when increasing the LoRA rank in the main paper. Higher learning rates effectively correspond to more aggressive post-training updates, leading to stronger in-domain adaptation at the cost of increased capability drift.

These results are not unexpected: stronger optimization pressure, whether induced by higher rank or larger learning rates, amplifies plasticity while degrading stability. Importantly, this trade-off is unlikely to be specific to LoRA and instead reflects a more general property of post-training:k increasing adaptation strength improves in-domain performance but exacerbates forgetting. As such, tuning optimization hyperparameters provides limited leverage for selectively mitigating forgetting without constraining learning.

\section{Limitations}
\label{sec:limitations}
\paragraph{Scope and abstraction of the taxonomy.}
CapTrack organizes model behavior into a fixed set of capability groups (CAN, WILL, HOW) to enable interpretable analysis of post-training effects. While this abstraction is useful for structuring evaluation, it does not capture all possible capabilities or their interactions. In particular, some behaviors may span multiple categories or emerge only in specific application contexts, and the taxonomy should therefore be understood as a practical approximation rather than a complete representation of model behavior.

\paragraph{Relative evaluation and interpretation of drift.}
CapTrack evaluates post-training effects through relative deviations from the corresponding base (OOB) model. This design improves comparability across models and benchmarks, but does not reflect absolute task performance. Not all observed deviations correspond to undesirable behavior: some changes reflect trade-offs (e.g., reduced verbosity or increased refusal rates) that may be beneficial depending on the application. Moreover, relative changes cannot verify any qualitative composition of behaviors and interpreting capability-level drift, requires context-dependent judgment.  Future work could, for example, investigate if numerically recovered capabilities (i.e. after applying DPO on top of IFT) reflect genuine behavioral restoration or superficial compensation mechanisms (e.g., padding, rigid response patterns).

\paragraph{LLM-as-a-judge evaluation.}
Several CapTrack metrics rely on LLM-as-a-judge evaluation for assessing properties such as reasoning quality, coverage, and faithfulness. While we validate the choice of judge model and use atomic evaluation protocols to improve robustness, such evaluations remain imperfect and may introduce noise or bias. Although our relative evaluation setup reduces sensitivity to systematic calibration differences, residual variance may still affect fine-grained comparisons.

\paragraph{Benchmark design and coverage.}
CapTrack builds on a curated set of existing benchmarks combined with targeted adaptations to probe specific capabilities. While this enables broad coverage in a sample-efficient manner, it necessarily inherits limitations of the underlying benchmarks, including potential dataset biases, limited coverage of certain behaviors, and sensitivity to prompt design. In addition, subsampling large benchmarks trades off evaluation cost against coverage, which may affect the stability of some estimates.

\paragraph{Comparison across model-families.}
Models across families have been subject to undisclosed and heterogeneous post-training procedures, potentially introducing confounding factors that cannot be fully controlled for. Furthermore, the use of relative deviation as primary metric yields a structural asymmetry that systematically makes higher-baseline models appear more stable than lower-baseline models despite the same absolute performance changes.

\paragraph{Experimental scope and generalization.}
Our empirical study is limited to a fixed set of model families, single domains (legal or medical), and post-training configurations (IFT and DPO under a standardized setup). While these choices reflect common real-world settings, the observed patterns may not fully generalize to other models, larger scales, alternative training procedures (e.g., RL-based alignment), or different data regimes (e.g. multi-domain)
. Further studies are required to assess the extent to which our findings hold in broader settings.

\paragraph{Use and interpretation.}
CapTrack is intended as a diagnostic framework for analyzing post-training effects rather than a standalone measure of model quality. Its metrics are most informative when used comparatively and in conjunction with application-specific evaluation.


\end{document}